\pdfoutput=1
\documentclass{article} %
\usepackage{iclr2027_conference,times}

\usepackage{amsmath,amsfonts,bm}

\def\eqref#1{equation~\ref{#1}}

\def\1{\bm{1}}

\DeclareMathAlphabet{\mathsfit}{\encodingdefault}{\sfdefault}{m}{sl}
\SetMathAlphabet{\mathsfit}{bold}{\encodingdefault}{\sfdefault}{bx}{n}

\usepackage{amsmath,amssymb}
\usepackage{graphicx,booktabs}
\usepackage{flafter}
\usepackage{algorithm}
\usepackage{algorithmic}
\usepackage{wrapfig}
\usepackage{enumitem}
\setlist[itemize]{leftmargin=*,nosep}

\usepackage{hyperref}
\usepackage{url}
\usepackage{xcolor}
\usepackage{colortbl} %
\usepackage{placeins} %
\usepackage{capt-of} %
\usepackage{tabularx} %
\newcolumntype{Y}{>{\centering\arraybackslash}X}
\newlength{\rqwidth}
\definecolor{ourrow}{HTML}{EEF2F7} %

\usepackage[normalem]{ulem} %
\newif\ifdraft
\drafttrue
\ifdraft
  \newcommand{\todo}[1]{{\color{red}\textbf{[TODO: #1]}}}
  \newcommand{\rev}[1]{{\leavevmode\color{black}#1}} %
  \newcommand{\del}[1]{{\color{red}\sout{#1}}}
\else
  \newcommand{\todo}[1]{\unskip}
  \newcommand{\rev}[1]{#1}
  \newcommand{\del}[1]{\unskip}
\fi

\newif\ifteaserabove
\teaserabovefalse
\newcommand{\teasercaption}{\rev{\textbf{Overview of RARA.} (Left) A policy is first pre-trained on diverse robot data. (Middle) To adapt it to a target task, RARA first retrieves pre-training demonstrations of a similar skill, such as removing a lid from a flask for the target task of opening a blender lid. It then aligns the target representation with their pre-trained representation, which it keeps as an anchor during fine-tuning. (Right) Across 13 simulated and real-world tasks, RARA improves the average ID and OOD success over fine-tuning by 11.6 points.}}

\title{RARA: Robust Policy Fine-Tuning through\\Retrieval-Anchored Representation\\Alignment for Robotic Manipulation}

\author{
  \textbf{Xuhui Kang, Sung-Wook Lee, Yan Zhuang, Tianle Zhong, Yen-Ling Kuo} \\
  University of Virginia \\
  \texttt{\{xuhui, dcs3zc, yyx3hd, fad3ew, ylkuo\}@virginia.edu}
}

\iclrfinalcopy %
\makeatletter
\def\@maketitle{\vbox{\hsize\textwidth
  {\raggedright\LARGE\sc \@title\par}
  \vskip 0.25in
  {\centering\@author\par}
  \vskip 0.3in minus 0.1in}}
\makeatother
\begin{document}
\maketitle
\lhead{Preprint} %

\ifteaserabove
\begin{center}
    \vspace{-2.0em}
    \includegraphics[width=\linewidth]{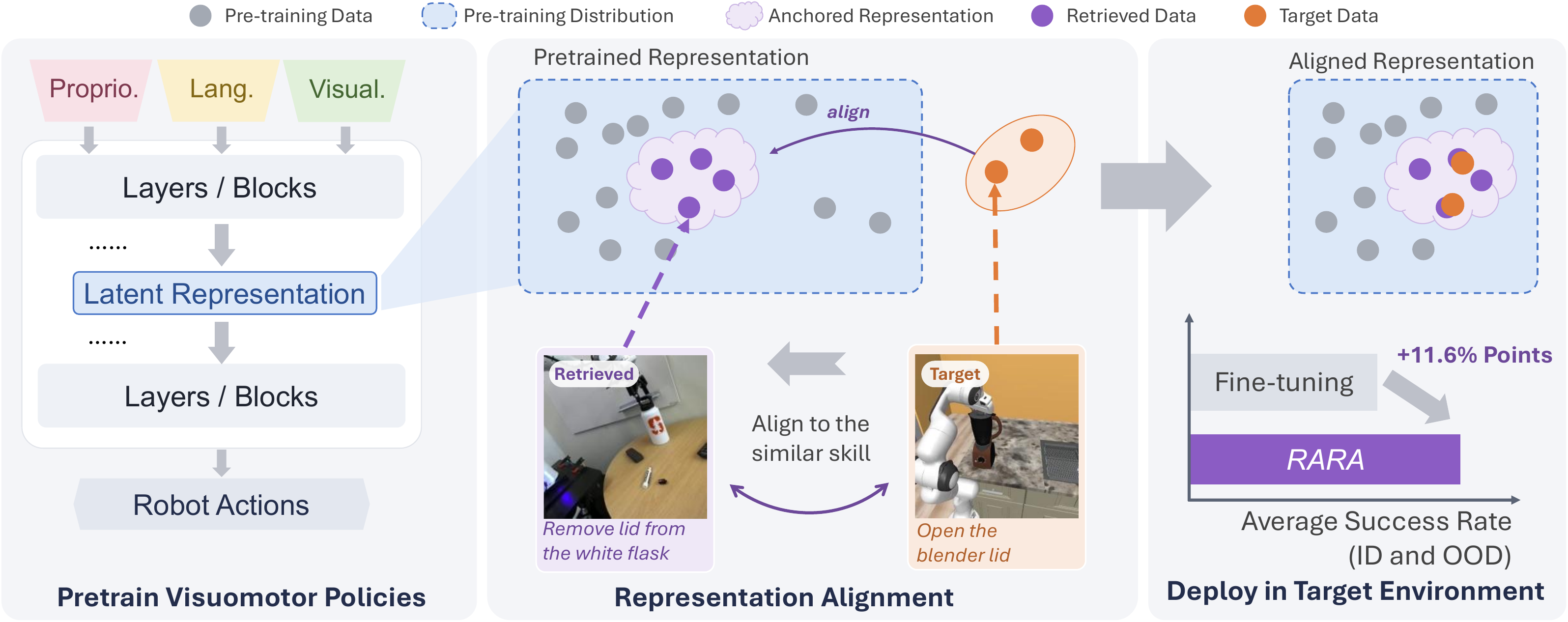}
    \vspace{-1.2em}
    \captionof{figure}{\teasercaption}
    \label{fig:teaser}
\end{center}
\vspace{-0.5em}
\fi

\begin{abstract}
    \rev{Pre-trained robot policies, such as} Vision-Language-Action (VLA) models\rev{,} provide a strong foundation for generalist robotic manipulation, but adapting them to new tasks with only a few demonstrations remains challenging. Directly fine-tuning a pre-trained policy often achieves strong in-distribution performance, yet it can distort the representations learned during pre-training and weaken the model's generalization ability under out-of-distribution shifts.
    We propose Retrieval-Anchored Representation Alignment (RARA), a two-stage adaptation method, to address this issue. First, RARA retrieves pre-training demonstrations with similar task intent and motion, and aligns the policy's representations of the target demonstrations with the pre-trained representations of the retrieved demonstrations. 
    Second, RARA fine-tunes the policy on the target demonstrations while limiting how much its representations of the retrieved demonstrations change during fine-tuning.
    By reducing the representation mismatch before action learning and preserving pre-trained knowledge during adaptation, RARA limits unnecessary representation drift during fine-tuning. Experiments with Diffusion Policy and GR00T on 13 tasks in RoboTwin, RoboCasa, and on a real robot show that RARA improves the average success of fine-tuning by 9.5 points in-distribution and 13.6 points out-of-distribution, and outperforms all compared baselines.
    \newline Project page: \href{https://shua-kang.github.io/RARA_robot/}{https://shua-kang.github.io/RARA\_robot/}
\end{abstract}

\section{Introduction}
\label{sec:intro}

\ifteaserabove\else
\begin{figure}[t]
    \centering
    \includegraphics[width=\linewidth]{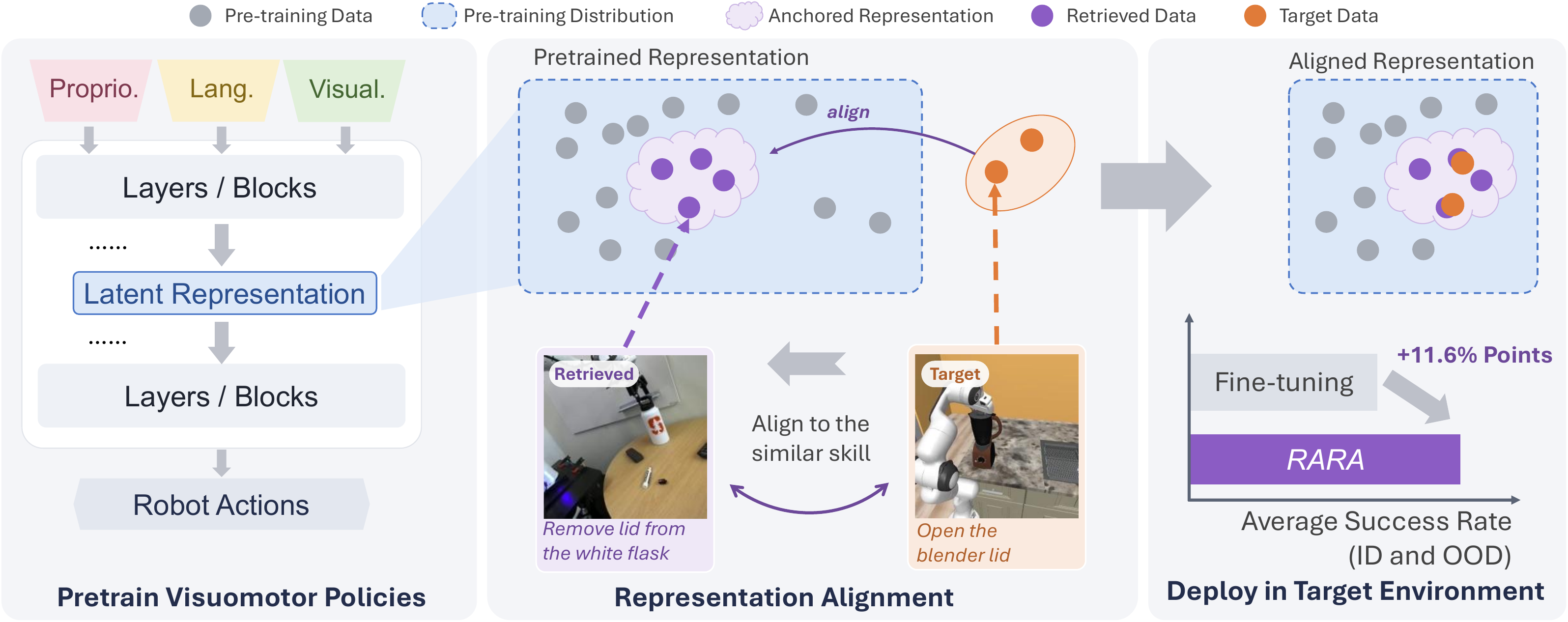}
    \vspace{-1.5em}
    \caption{\teasercaption}
    \label{fig:teaser}
    \vspace{-0.5em}
\end{figure}
\fi

Pre-training on large-scale robot demonstrations followed by fine-tuning on a small target dataset has become a standard practice in learning robot manipulation policies~\citep{o2024open,khazatsky2024droid,black2024pi_0,bjorck2025gr00t}.
By leveraging supervision from large and diverse robot demonstration datasets, pre-training provides useful initial representations that help adaptation to a new task.
However, fine-tuning on limited target data can still overfit the policy to the narrow distribution seen during adaptation:
A policy may succeed on the target task under the in-distribution conditions, but fail when the camera viewpoint, background, or object properties change~\citep{fei25libero-plus}.
This vulnerability arises because standard fine-tuning on limited data can severely distort the representations learned during pre-training~\citep{kumar2022finetuning,kachaev_dont_2025,yadav2025robust}. 
Therefore, \emph{robust policy fine-tuning}, which requires learning a target task while maintaining generalization to out-of-distribution shifts, remains a critical challenge when adapting pre-trained robotic policies to new tasks.

Existing methods have explored several ways to preserve generalizable representations when adapting policies to new tasks.
One line of work constrains the adapted policy toward its pre-trained initialization, for example by regularizing fine-tuned weights toward the pre-trained weights~\citep{huang2025maps}, merging pre-trained and fine-tuned weights~\citep{yadav2025robust}, or aligning policy representations with those of a frozen pre-trained visual encoder~\citep{kachaev_dont_2025}.
Another line of work distills knowledge from multiple pre-trained models into a shared representation for downstream robot learning~\citep{shang2024theia}.
While these approaches preserve broad pre-trained knowledge, they do not provide a principled way to focus alignment to pre-training demonstrations that are most relevant to the target task.

We propose \textbf{R}etrieval-\textbf{A}nchored \textbf{R}epresentation \textbf{A}lignment (\textbf{RARA}), which uses task-relevant pre-training demonstrations as anchors for fine-tuning (Fig.~\ref{fig:teaser}). 
RARA retrieves relevant demonstrations from the pre-training data by matching both task instructions and robot motions to the target demonstrations. For example, demonstrations of removing a pot lid may be retrieved for target tasks like opening a blender and remove a bottle cap.
The retrieved demonstrations support adaptation in two complementary ways.
Before fine-tuning, RARA aligns every target frame with the retrieved frames whose upcoming actions are most similar, so each frame is matched to the same stage of the motion in the retrieved demonstrations (e.g., a frame in which the gripper closes around the blender lid is matched to frames in which it closes around a pot lid).
The pre-trained policy has seen these stages under many visual conditions, so alignment lets the target frames reuse this robust representation instead of fitting the target demonstrations from scratch.
During fine-tuning, an anchor loss keeps the representations of the retrieved demonstrations close to their pre-trained values, so the small target dataset does not overwrite this robust structure.
Unlike a standard drift regularizer, the anchor acts only on the retrieved demonstrations. It therefore does not pull the target representation back, and it preserves only the pre-trained structure that is related to the target task.

We fine-tune a pre-trained Diffusion Policy on four RoboTwin tasks, and a DROID-pre-trained GR00T policy on five RoboCasa tasks and four real-robot tasks. We evaluate each policy in the conditions covered by its demonstrations (in-distribution, ID) and under new camera poses, backgrounds, distractors, objects or object positions (out-of-distribution, OOD). Across these 13 tasks, RARA improves the average success rates of fine-tuned policies by $9.5$ \% points in ID and $13.6$ \% points in OOD settings, consistently outperforming the baseline methods.

Our contributions are threefold: (1) we introduce RARA, a framework for robust policy fine-tuning that uses task-relevant pre-training experience to guide representation learning; (2) we evaluate RARA across different policies, simulated environments, and real-world manipulation tasks, measuring performance of target tasks and robustness to unseen conditions; and (3) we conduct extensive studies that separate initial alignment from representation constraints during fine-tuning, and examine how the choice of source data and network layer affects adaptation.

\section{Related Work}
\label{sec:related_work}

\noindent\textbf{Robust fine-tuning.} The goal of robust policy fine-tuning is to learn a downstream task without erasing the generalization acquired during pre-training~\citep{yadav2025robust, grover2025enhancing, hancock2025actions, yang2026visionlanguageaction, driess2025knowledge}. In transfer learning, \citet{kumar2022finetuning} show that full fine-tuning can distort pre-trained features and hurt OOD robustness, motivating staged or constrained updates. Other methods keep the fine-tuned features close to those of the pre-trained model~\citep{mukhoti2024ldifs}, or use auxiliary data with rich semantics as anchors that preserve OOD generalization~\citep{han2024arf}. In robot learning, one line of work freezes parts of the model and adds lightweight trainable modules, such as adapters or task-specific heads, to preserve existing capabilities while specializing to new tasks~\citep{liang2022transformer,sharma2023lossless,liu2024tail,xu2023hyper}. 
Another line preserves the capabilities of the pre-trained policy more directly. Weight-space methods merge the pre-trained and fine-tuned weights~\citep{yadav2025robust,dey2025revla} or constrain how far the weights move~\citep{huang2025maps}. Other methods keep a frozen copy of the pre-trained model next to the adapted one, as a frozen encoder with a trainable branch~\citep{grover2025enhancing}, a frozen prior in an ensemble~\citep{bagatella2024active}, or a frozen expert that guides an adaptation expert~\citep{guo_priorvla_2026}. Representation-space methods align the features of the adapted policy with those of a frozen copy or teacher on the target data~\citep{dalal2026anchor,lin2026fibertune}.
These methods preserve the pre-trained policy through its architecture, its weights, or the representation of the target data. RARA instead anchors the representation of retrieved pre-training demonstrations: 
it aligns the target representation with the pre-trained representation of these demonstrations, and then limits how much it changes during fine-tuning,
without restricting the architecture or which weights are updated.

\noindent\textbf{Domain adaptation.} Robot policies often face domain gaps between simulation and the real world, different embodiments, or pre-training and downstream deployment environments. Domain adaptation methods therefore seek task-relevant representations or data that transfer across such shifts. For VLA models, \citet{kachaev_dont_2025} identify degradation of visual representations after action fine-tuning and preserve visual grounding through alignment with a frozen vision teacher, while \citet{luo_look_2026} show that action generation can become insufficiently sensitive to relevant visual tokens in deeper layers, and \citet{kim2026rscl} regularize VLA representations with a contrastive loss whose positives are defined by robot states. In sim-and-real co-training, \citet{maddukuri2025sim} show that mixing simulated and real data improves real-world success, \citet{cheng_generalizable_2025} align joint observation-action distributions with optimal-transport-inspired objectives,  
while \citet{lei_mechanistic_nodate} combine adversarial alignment with explicit domain conditioning, which keeps the sim and real domains distinguishable in the learned representation space.
Data selection offers a complementary route: Behavior Retrieval and FlowRetrieval retrieve prior data that are similar to the target demonstrations in state-action or motion space~\citep{du2023behavior,lin2024flowretrieval}, and \citet{dass2025datamil} select useful data from a pre-training dataset for a target imitation-learning task, improving the adapted policy with task-relevant source examples. 
Similarly, RARA selects task-relevant data from the pre-training dataset, but it uses the selected demonstrations as anchors for representation alignment rather than as additional training data: it aligns the target representation with that of the selected demonstrations and limits how much their representation changes during fine-tuning.

\section{Problem Definition}
\label{sec:problem_setting}

We consider adapting a visuomotor policy $\pi_{\theta_{\mathrm{pre}}}$, pre-trained on a large and diverse set of robot demonstrations $D_{\mathrm{pre}}$, to a downstream target task with a limited target demonstration set $D_{\mathrm{tgt}}=\{(o_i,a_i)\}_{i=1}^{N}$.
Here $o_i$ denotes the full observation available to the policy, including visual inputs and any other language or state information, and $a_i$ denotes the demonstrated action. Standard target-only fine-tuning updates the policy by behavior cloning,
\begin{equation}
    \theta_{\mathrm{ft}}
    = \arg\min_{\theta}
    \mathbb{E}_{(o,a) \sim D_{\mathrm{tgt}}}
    \left[\mathcal{L}_{BC}\big(\pi_{\theta}(o), a\big)\right],
    \label{eq:main_full_ft}
\end{equation}
where $\mathcal{L}_{BC}$ is the behavior cloning loss. While this objective directly optimizes performance on the in-distribution (ID) target data, a robust policy must also remain reliable under out-of-distribution (OOD) shifts after adaptation. Therefore, robust fine-tuning aims to learn a policy that achieves high expected task success under both ID and OOD evaluation distributions:
\begin{equation}
    \max_{\theta} \; J_{\mathrm{ID}}(\pi_{\theta}) + J_{\mathrm{OOD}}(\pi_{\theta}),
    \qquad
    J_{\mathcal{E}}(\pi_{\theta})
    = \mathbb{E}_{\tau \sim p_{\mathcal{E}}(\pi_{\theta})}\left[S(\tau)\right],
    \quad \mathcal{E}\in\{\mathrm{ID},\mathrm{OOD}\},
    \label{eq:main_goal}
\end{equation}
where $S(\tau)\in\{0,1\}$ indicates task success for a trajectory $\tau$, and $p_{\mathcal{E}}$ denotes the trajectory distribution induced by the policy $\pi_{\theta}$ in evaluation setting $\mathcal{E}$.

\vspace{-0.5em}
\section{RARA: Retrieval-Anchored Representation Alignment}
\label{sec:method}

RARA adapts a pre-trained policy to a target task in two stages. It first aligns the target representation to the pre-trained representation to obtain a good initial representation, and then fine-tunes the policy on the target task while constraining representation shifts of pretraining tasks. 
Both stages use demonstrations retrieved from the pre-training data (\S\ref{sec:retrieval}). In Stage~1, the target representation is aligned with their pre-trained representation, and in both stages their own representation is anchored.

\begin{figure}[t]
    \centering
    \includegraphics[width=\linewidth]{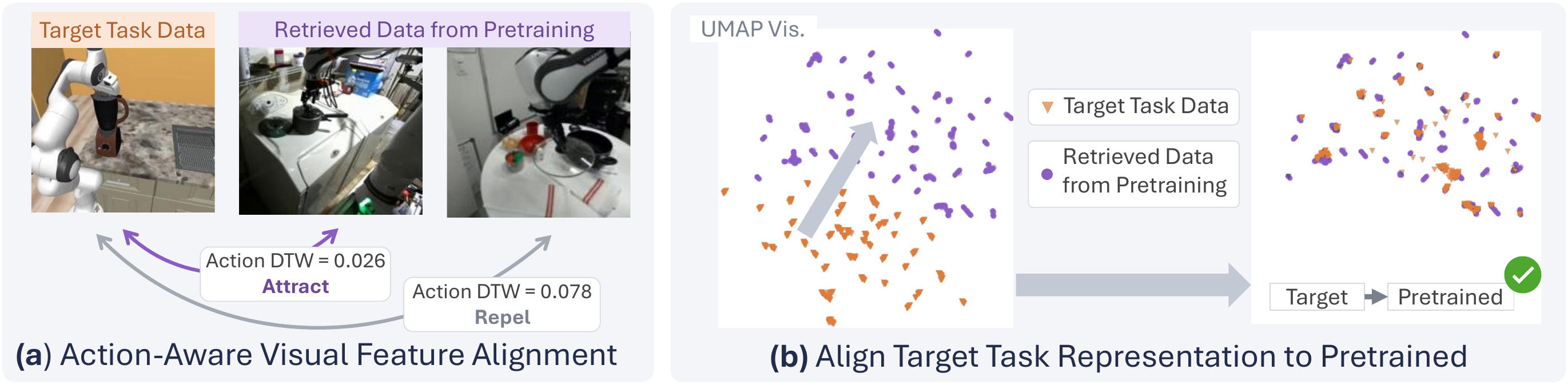}
    \vspace{-1.5em}
    \caption{\textbf{Stage~1 of RARA.} (a)~For a target frame, frames of the retrieved demonstrations with similar action chunks (small DTW distance) are pulled closer, and frames with different action chunks are pushed away. (b)~UMAP of the representations before (left) and after (right) Stage~1: the target representations (orange) move toward those of the retrieved pre-training data (purple).}
    \label{fig:stage1_align}
    \vspace{-0.3em}
\end{figure}

\paragraph{Stage 1: Aligning to the pre-trained representation.}
Before fully fine-tuning the policy, Stage~1 obtains an initial representation of the target demonstrations by aligning it to the pre-trained representation. Specifically, it aligns the representation of each target frame with the pre-trained representation of retrieved frames at the matched steps of the task (Fig.~\ref{fig:stage1_align}). An anchor loss limits how much this alignment changes the representation of the retrieved demonstrations. Only the layers that produce the representation and a projector that maps pooled representation into an embedding space are trained in this stage.

\paragraph{Stage 2: Fine-tuning on the target task.}
Stage~2 fine-tunes the policy on the target demonstrations with its behavior cloning loss. The anchor loss stays active, so fine-tuning preserves the pre-trained representation on the retrieved demonstrations while the policy adapts to the target task. The projector is discarded, and only the fine-tuned policy is used at inference.

Formally, the two stages solve
\begin{align}
    \text{Stage 1:}\quad & \min_{\theta,\,\psi}\; \mathcal{L}_{\mathrm{contrastive}} + \lambda_1\,\mathcal{L}_{\mathrm{anchor}},
    \label{eq:stage1_objective}\\
    \text{Stage 2:}\quad & \min_{\theta}\; \mathcal{L}_{\mathrm{BC}} + \lambda_2\,\mathcal{L}_{\mathrm{anchor}},
    \label{eq:anchored_finetuning}
\end{align}
where $\mathcal{L}_{\mathrm{BC}}$ is the behavior-cloning loss on the target demonstrations (Eq.~\ref{eq:main_full_ft}), $\mathcal{L}_{\mathrm{contrastive}}$ aligns target and retrieved representations, $\mathcal{L}_{\mathrm{anchor}}$ limits the change of the retrieved representations, and $\psi$ are the projector parameters. In Stage~1, $\theta$ includes only the layers that produce the anchored representation; in Stage~2, the trainable parameters follow the standard fine-tuning recipe of the policy. The rest of this section describes retrieval (\S\ref{sec:retrieval}) and the two losses (\S\ref{sec:alignment}).

\subsection{Retrieving Relevant Pre-training Demonstrations}
\label{sec:retrieval}
\begin{wrapfigure}{R}{0.65\textwidth}
    \vspace{-1.2em}
    \centering
    \includegraphics[width=\linewidth]{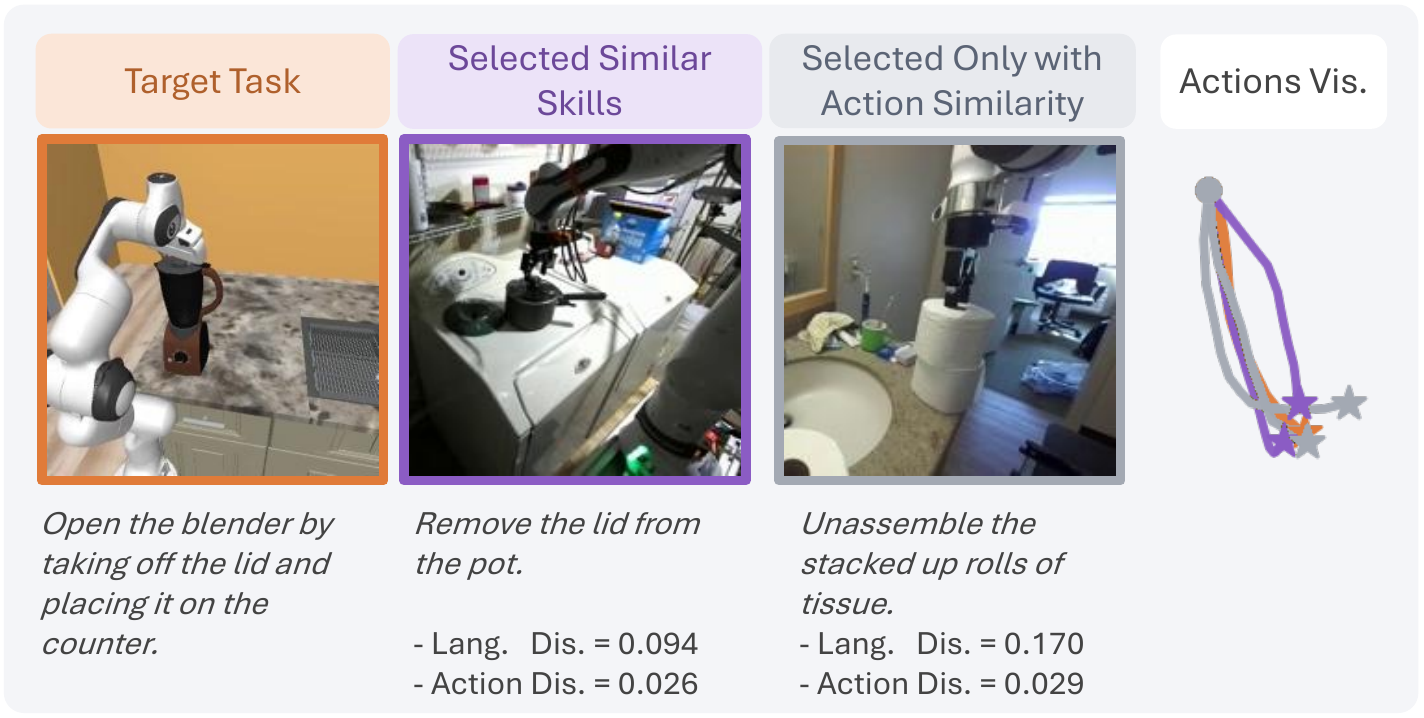}
    \vspace{-1.6em}
    \caption{Pre-training demonstration retrieval for the target task: Open Blender Lid. Selecting by action similarity alone can return similar action chunks from a different task (grey); retrieving by the task instruction returns the same kind of task (purple).}
    \label{fig:retrieval_example}
    \vspace{-1.0em}
\end{wrapfigure}
RARA uses pre-training demonstrations relevant to the target task as a reference for the policy's representation update. We define relevance at two levels. 
A relevant \emph{demonstration} performs the same type of task as the target, 
even though the objects and scenes may differ; we find such demonstrations by instruction similarity.
A relevant \emph{frame} is at the matched step of the task as a target frame; 
we find such frames by comparing action chunks (i.e., short sequences of consecutive actions).
Both levels are necessary. Fig.~\ref{fig:retrieval_example} shows an example. Removing a pot lid and unstacking tissue rolls both lift an object upward, so their action chunks are similar even though the tasks are different. Action similarity alone would pair a target frame with a frame of a different task.

\paragraph{Retrieving demonstrations.}
We embed task instructions with a frozen CLIP text encoder~\citep{radford2021learning}. Because a target task can have several instructions, we average their normalized embeddings. We then rank the pre-training demonstrations by the cosine similarity between their instruction embedding and this average, and keep the top $K$ demonstrations. 

\paragraph{Pairing frames by action chunks.}
We split every demonstration in $D_{\mathrm{ret}}$ and $D_{\mathrm{tgt}}$ into overlapping action chunks $A_t=[a_t,\ldots,a_{t+H-1}]$ and represent each chunk by the observation $o_t$ at its first step. Let $\delta(A,A')$ be the dynamic-time-warping (DTW) distance between two action chunks, which matches similar actions executed at different speeds. Defining positives by action similarity follows CLASS~\citep{lee2025class}; here the pairs connect target frames with retrieved pre-training frames. 
For a target frame $o_i^{\mathrm{tgt}}$ with action chunk $A_i^{\mathrm{tgt}}$, we compute the DTW distance $\delta(A, A_i^{\mathrm{tgt}})$ from every retrieved chunk $A$ to $A_i^{\mathrm{tgt}}$.
The positives $\mathcal{P}(i)$ are the frames of the $m_{+}$ retrieved chunks with the smallest distance.
The negatives $\mathcal{N}(i)$ are frames sampled from the retrieved chunks whose distance is above the median of these distances.
Using the median as the threshold is a practical choice that keeps the negatives clearly different from the target chunk while leaving many chunks to sample from.
RARA uses the actions of the retrieved demonstrations only to form these pairs, not as behavior-cloning targets.

\subsection{Aligning an Intermediate Representation}
\label{sec:alignment}

\rev{Let $h_{\theta}^{\ell}(o)$ denote the representation at layer $\ell$ of policy $\pi_{\theta}$, and let $\bar h^{\ell}(o)=h_{\theta_{\mathrm{pre}}}^{\ell}(o)$ be the corresponding representation from a frozen copy of the pre-trained policy.} The observation $o$ includes the policy's image, language and state inputs, as applicable. 

\paragraph{Choosing the anchored layer.}
We call $\ell$ the \emph{anchored layer} and $h_{\theta}^{\ell}$ the \emph{anchored representation}. Both the contrastive loss and the anchor loss act on this representation. The anchored layer can be any layer of the policy, from an early perception layer to a layer inside the action network, so RARA does not depend on a particular policy architecture.
In practice, the target data usually differ from the pre-training data mainly in visual appearance. In this case, any layer that follows the visual input can serve as the anchored layer.
We keep $\ell$ general below; \S\ref{sec:exp_setup} gives the layer used in our experiments, and RQ3 (\S\ref{sec:exp_layers}) shows that the gain is not tied to one layer.

\paragraph{Matching target representations to source references.}
For a target observation $o_i^{\mathrm{tgt}}$, let $\mathcal{P}(i)$ and $\mathcal{N}(i)$ be its positive and negative frames from $D_{\mathrm{ret}}$ (\S\ref{sec:retrieval}).
We compare representations in a contrastive embedding space. 
The representation is pooled over tokens or cameras as needed and mapped to an $\ell_2$-normalized embedding by a trainable two-layer MLP projector $g_{\psi}$. Following common practice in contrastive learning, we compute the loss after this projector rather than directly on the representation, which has been shown to improve the learned representation and to reduce its dimensional collapse~\citep{chen2020simple,jing2022understanding} (see Appx.~\ref{app:projector} for details).
The embedding of the target frame is computed with the adapting policy, $z_i^{\mathrm{tgt}}=g_{\psi}(h_{\theta}^{\ell}(o_i^{\mathrm{tgt}}))$, and the embedding of a retrieved frame $o_j$ is computed with the frozen pre-trained policy, $\bar z_j=g_{\psi}(\bar h^{\ell}(o_j))$. A multi-positive contrastive loss~\citep{oord2018representation,khosla2020supervised,miech2020end} then aligns each target embedding with the embeddings of its positive frames (Fig.~\ref{fig:stage1_align}):
\begin{equation}
    \mathcal{L}_{\mathrm{contrastive}}
    = -\frac{1}{B}\sum_{i=1}^{B}\log
    \frac{\sum_{p\in\mathcal{P}(i)}\exp\big(\mathrm{sim}(z_i^{\mathrm{tgt}},\bar z_p)/\tau\big)}
    {\sum_{p\in\mathcal{P}(i)}\exp\big(\mathrm{sim}(z_i^{\mathrm{tgt}},\bar z_p)/\tau\big)
    +\sum_{n\in\mathcal{N}(i)}\exp\big(\mathrm{sim}(z_i^{\mathrm{tgt}},\bar z_n)/\tau\big)},
    \label{eq:contrastive_loss}
\end{equation}
where $B$ is the target batch size, $\mathrm{sim}(u,v)=u^{\top}v/(\|u\|\,\|v\|)$ is cosine similarity, and $\tau$ is the temperature. The pre-trained policy is frozen, so the representations $\bar h^{\ell}$ of the retrieved frames stay fixed and serve as the reference for the anchor loss below; only their embeddings $\bar z_j$ change as the projector is trained. The projector is used only for the contrastive loss and is discarded after Stage~1.

\paragraph{Anchoring source representations.}
\rev{Updating the policy to represent the target also changes how it represents source observations.} We constrain these changes with
\begin{equation}
    \mathcal{L}_{\mathrm{anchor}}^{\ell}(\theta)
    = \mathbb{E}_{o\sim D_{\mathrm{ret}}}
    \left[d\!\left(h_{\theta}^{\ell}(o),\bar h^{\ell}(o)\right)\right],
    \label{eq:anchor_loss}
\end{equation}
where $d$ measures the discrepancy between the two representations (see Appx.~\ref{app:implementation_details} for details). The loss compares the adapting and the pre-trained policy on the same frame of a retrieved demonstration but uses no action labels. It is computed only on the retrieved demonstrations, so it leaves the target representation free to change under the contrastive loss and, later, the task loss; it only limits how much these changes also alter the representation of the retrieved demonstrations. Together with the contrastive loss, it forms the Stage-1 objective in Eq.~\ref{eq:stage1_objective}, and it stays active in Stage~2 (Eq.~\ref{eq:anchored_finetuning}).

\section{Experiments}
\label{sec:experiments}

We evaluate whether RARA learns a target task while generalizing beyond the conditions covered by its demonstrations, by fine-tuning two pre-trained policies on 13 tasks in simulation and on a real robot (\S\ref{sec:exp_setup}). We aim to answer the following four research questions:
\begin{itemize}
    \item \textbf{RQ1:} How does RARA change the representation compared with standard fine-tuning? (\S\ref{sec:exp_alignment})
    \item \textbf{RQ2:} Does RARA improve ID and OOD success over fine-tuning and prior methods? (\S\ref{sec:exp_main})
    \item \textbf{RQ3:} Does the gain depend on the choice of the anchored layer? (\S\ref{sec:exp_layers})
    \item \textbf{RQ4:} Which components and data contribute to the gain? (\S\ref{sec:exp_ablation})
\end{itemize}

\subsection{Setup}
\label{sec:exp_setup}

\begin{figure}[t]
    \centering
    \includegraphics[width=\linewidth]{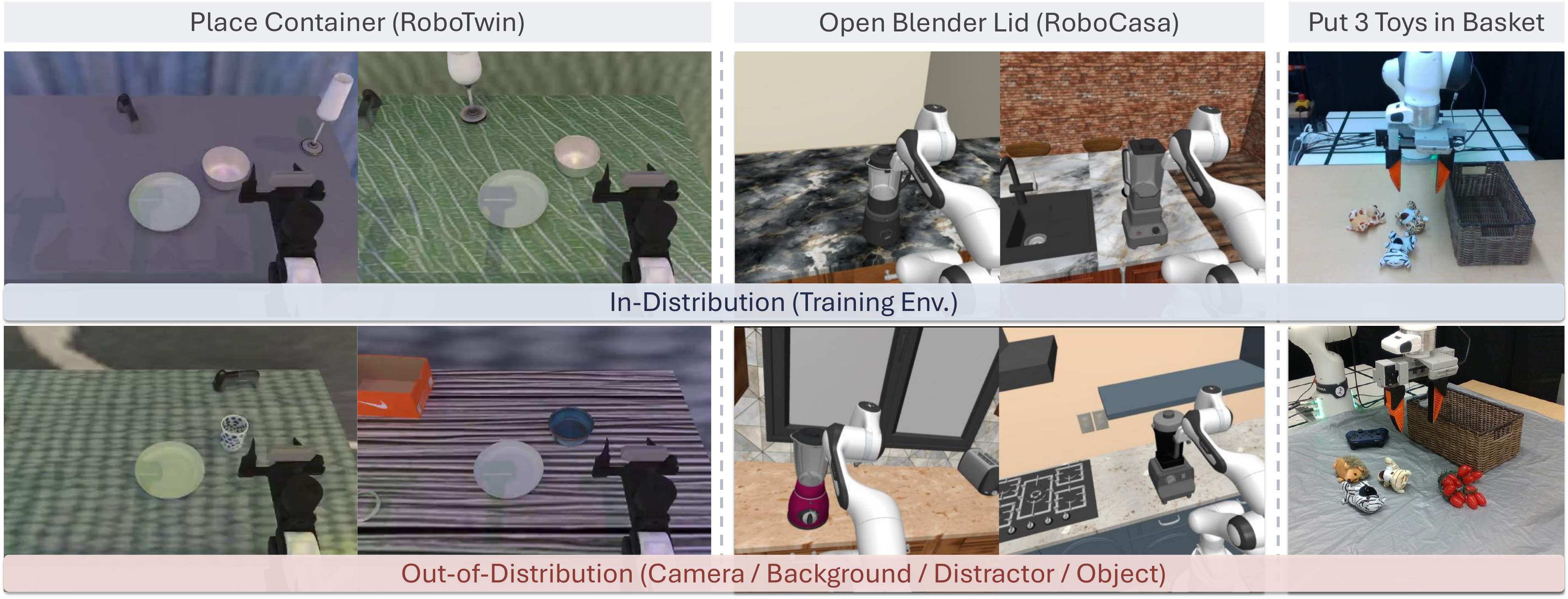}
    \vspace{-1.5em}
    \caption{ID (top) and OOD (bottom) scenes for one \rev{RoboTwin,} one RoboCasa and one real-robot task.}
    \label{fig:train_eval_envs}
    \vspace{-1.0em}
\end{figure}

We evaluate RARA in two simulated environments and on a real robot (Fig.~\ref{fig:train_eval_envs}). In each setting, a pre-trained policy is fine-tuned on a small set of target demonstrations and evaluated both in-distribution (ID) and out-of-distribution (OOD). See Appx.~\ref{app:evaluation_envs} for the full details of setup.
\begin{itemize}
    \item \textbf{RoboTwin~2.0}~\citep{chen2025robotwin2}. We use four single-arm target tasks, such as Place Container and Move Pill Bottle. ID uses the narrow layout of the target demonstrations; OOD widens the object layout, the backgrounds and the distractors. For pre-training, we collect about 2k demonstrations of various skills in RoboTwin, such as pick-and-place and pressing a bell, holding out the four target tasks. The pre-training data cover a larger workspace and more distractors, objects and background textures than the target demonstrations. Each target task has 50 demonstrations.
    \item \textbf{RoboCasa}~\citep{robocasa365}. We use five target tasks, such as Open Blender Lid and Close Fridge Drawer. ID uses the demonstration kitchen with randomized object poses; OOD changes the camera pose, background, distractors and object instances. Pre-training uses DROID~\citep{khazatsky2024droid}, which contains 76k demonstrations, and each target task has 100 demonstrations collected in one kitchen.
    \item \textbf{Real Franka robot}. We use four target tasks, such as Remove Pot Lid and Load Bread in Toaster. ID and OOD are defined as in RoboCasa. Pre-training uses DROID, and each target task has 50 to 80 demonstrations collected by VR teleoperation.
\end{itemize}

\paragraph{Policies.}
We evaluate two common visuomotor policies:
\begin{itemize}
    \item \textbf{Diffusion Policy}~\citep{chi2023diffusion}, a widely used visuomotor policy, which we pre-train on the RoboTwin pre-training data.
    \item \textbf{GR00T~N1.7}~\citep{bjorck2025gr00t}, a large vision-language-action (VLA) model pre-trained on DROID, which we use for RoboCasa and the real robot.
\end{itemize}
Since the target tasks differ from the pre-training data mainly in visual appearance, the anchored layer of both policies is the output of the layers that encode the visual input. 
All methods in a setting use the same optimizer settings and fine-tuning budget, and RARA additionally runs a short Stage~1 before fine-tuning.
See Appx.~\ref{app:implementation_details} for details.

\paragraph{Evaluation Protocol.}
We evaluate each policy on 150 rollouts per task and setting in RoboTwin, 50 in RoboCasa and 10 trials on the real robot, and repeat each experiment three times (see Appx.~\ref{app:evaluation_envs} for details). On the real robot, all methods are tested on the same recorded scene randomizations. We report the mean and standard deviation of the success rate.

\paragraph{Baselines.}
We compare RARA with five baselines. They cover the two common ways to reuse pre-training knowledge during fine-tuning: training on pre-training data (Co-FT and STRAP) and keeping the policy close to the pre-trained one (RETAIN and BlindVLA). \textbf{FT} fine-tunes the policy on the target demonstrations only. \textbf{Co-FT} also trains on the retrieved pre-training demonstrations $D_{\mathrm{ret}}$ (\S\ref{sec:retrieval}) with the behavior-cloning loss. \textbf{STRAP}~\citep{memmel2024strap} instead retrieves pre-training sub-trajectories that match the target demonstrations under dynamic time warping, and co-trains on them. \textbf{RETAIN}~\citep{yadav2025robust} merges the weights of the pre-trained and fine-tuned policies. \textbf{BlindVLA}~\citep{kachaev_dont_2025} aligns the policy's visual representations with those of a frozen pre-trained visual encoder during fine-tuning. See Appx.~\ref{app:robotwin_baselines} for details.

\subsection{RQ1: How Does RARA Change the Representation?}
\label{sec:exp_alignment}

\begin{table}[t]
\centering
\caption{Representation metrics at the anchored layer on RoboTwin (mean over five pre-training seeds). Left: computed on held-out pre-training frames; higher values mean that more of the pre-trained structure remains. Right: feature distortion on held-out frames of the target task in the ID and OOD conditions; lower values mean smaller changes from the pre-trained policy $P$. Bold marks the best adapted value in each column. ``After Stage~1'' is the policy before Stage-2 fine-tuning.}
\label{tab:rq1}
\footnotesize
\setlength{\tabcolsep}{4pt}
\renewcommand{\arraystretch}{1.08}
\begin{tabular}{@{}l*{10}{c}@{}}
\toprule
     & \multicolumn{6}{c}{Pre-training feature retention}
 & \multicolumn{4}{c}{Target-data feature distortion} \\
\cmidrule(lr){2-7}\cmidrule(l){8-11}
 & \multicolumn{3}{c}{Move pill bottle} & \multicolumn{3}{c}{Place container}
 & \multicolumn{2}{c}{Move pill bottle} & \multicolumn{2}{c}{Place container} \\
\cmidrule(lr){2-4}\cmidrule(lr){5-7}\cmidrule(lr){8-9}\cmidrule(l){10-11}
Recipe & Rank $\uparrow$ & CKA $\uparrow$ & $R^2$ $\uparrow$
       & Rank $\uparrow$ & CKA $\uparrow$ & $R^2$ $\uparrow$
       & ID $\downarrow$ & OOD $\downarrow$ & ID $\downarrow$ & OOD $\downarrow$ \\
\midrule
Pre-trained $P$ & 58.8 & 1.00 & 0.71 & 58.8 & 1.00 & 0.71 & 0.00 & 0.00 & 0.00 & 0.00 \\
FT & 50.2 & 0.68 & 0.67 & 53.1 & 0.87 & 0.64 & 0.50 & 0.49 & 0.47 & 0.48 \\
Co-FT & 51.4 & \textbf{0.93} & 0.72 & 49.1 & \textbf{0.94} & 0.72 & 0.57 & 0.57 & 0.62 & 0.63 \\
\rowcolor{ourrow} RARA, after Stage~1 & \textbf{73.2} & 0.70 & \textbf{0.74} & \textbf{68.6} & 0.76 & \textbf{0.75} & 0.48 & 0.46 & 0.56 & 0.53 \\
\rowcolor{ourrow} RARA & 54.5 & 0.92 & 0.71 & 55.7 & 0.91 & 0.70 & \textbf{0.23} & \textbf{0.23} & \textbf{0.31} & \textbf{0.33} \\
\bottomrule
\end{tabular}
\end{table}

\begin{table}[t]
\centering
\caption{Simulation results (success rate, \%, mean\,$\pm$\,std). RoboTwin uses the Diffusion Policy; RoboCasa uses GR00T~N1.7 pre-trained on DROID. Bold: best; underline: second best.}
\label{tab:main}
\scriptsize
\setlength{\tabcolsep}{2pt}
\setlength{\rqwidth}{\linewidth}
\begin{tabularx}{\rqwidth}{@{}l*{10}{Y}@{}}
\toprule
\textbf{RoboTwin} & \multicolumn{2}{c}{\shortstack{Place\\Container}} & \multicolumn{2}{c}{\shortstack{Move Pill\\Bottle}} & \multicolumn{2}{c}{\shortstack{Click\\Alarm Clock}} & \multicolumn{2}{c}{\shortstack{Shake\\Bottle}} & \multicolumn{2}{c}{\shortstack{Task\\average}} \\
\cmidrule(lr){2-3}\cmidrule(lr){4-5}\cmidrule(lr){6-7}\cmidrule(lr){8-9}\cmidrule(lr){10-11}
Method & ID & OOD & ID & OOD & ID & OOD & ID & OOD & ID & OOD \\
\midrule
FT & \underline{82.0}{\fontsize{4}{5}\selectfont$\,\pm$0.7} & \underline{60.7}{\fontsize{4}{5}\selectfont$\,\pm$2.4} & 32.9{\fontsize{4}{5}\selectfont$\,\pm$3.7} & 9.6{\fontsize{4}{5}\selectfont$\,\pm$2.3} & 80.4{\fontsize{4}{5}\selectfont$\,\pm$6.3} & 52.4{\fontsize{4}{5}\selectfont$\,\pm$4.1} & 74.2{\fontsize{4}{5}\selectfont$\,\pm$2.1} & 66.4{\fontsize{4}{5}\selectfont$\,\pm$5.7} & 67.4{\fontsize{4}{5}\selectfont$\,\pm$1.4} & 47.3{\fontsize{4}{5}\selectfont$\,\pm$2.4} \\  %
Co-FT & 62.7{\fontsize{4}{5}\selectfont$\,\pm$5.8} & 43.3{\fontsize{4}{5}\selectfont$\,\pm$3.3} & 32.4{\fontsize{4}{5}\selectfont$\,\pm$8.1} & 10.0{\fontsize{4}{5}\selectfont$\,\pm$2.4} & \textbf{81.8}{\fontsize{4}{5}\selectfont$\,\pm$3.4} & 61.3{\fontsize{4}{5}\selectfont$\,\pm$1.8} & 80.7{\fontsize{4}{5}\selectfont$\,\pm$6.8} & \underline{75.6}{\fontsize{4}{5}\selectfont$\,\pm$5.7} & 64.4{\fontsize{4}{5}\selectfont$\,\pm$2.8} & 47.6{\fontsize{4}{5}\selectfont$\,\pm$1.3} \\  %
STRAP & 81.1{\fontsize{4}{5}\selectfont$\,\pm$3.8} & 59.3{\fontsize{4}{5}\selectfont$\,\pm$2.4} & 44.4{\fontsize{4}{5}\selectfont$\,\pm$5.1} & 13.8{\fontsize{4}{5}\selectfont$\,\pm$1.4} & \underline{81.3}{\fontsize{4}{5}\selectfont$\,\pm$1.3} & \underline{61.6}{\fontsize{4}{5}\selectfont$\,\pm$1.7} & \underline{83.6}{\fontsize{4}{5}\selectfont$\,\pm$7.3} & 74.9{\fontsize{4}{5}\selectfont$\,\pm$7.7} & \underline{72.6}{\fontsize{4}{5}\selectfont$\,\pm$2.0} & \underline{52.4}{\fontsize{4}{5}\selectfont$\,\pm$2.3} \\  %
RETAIN & 55.3{\fontsize{4}{5}\selectfont$\,\pm$10.1} & 40.7{\fontsize{4}{5}\selectfont$\,\pm$4.2} & 29.3{\fontsize{4}{5}\selectfont$\,\pm$6.1} & 13.3{\fontsize{4}{5}\selectfont$\,\pm$3.1} & 53.3{\fontsize{4}{5}\selectfont$\,\pm$6.1} & 36.0{\fontsize{4}{5}\selectfont$\,\pm$4.0} & 50.0{\fontsize{4}{5}\selectfont$\,\pm$6.0} & 46.7{\fontsize{4}{5}\selectfont$\,\pm$6.4} & 47.0{\fontsize{4}{5}\selectfont$\,\pm$6.7} & 34.2{\fontsize{4}{5}\selectfont$\,\pm$2.3} \\  %
BlindVLA & 69.3{\fontsize{4}{5}\selectfont$\,\pm$3.1} & 52.9{\fontsize{4}{5}\selectfont$\,\pm$2.0} & \underline{47.6}{\fontsize{4}{5}\selectfont$\,\pm$1.4} & \underline{15.8}{\fontsize{4}{5}\selectfont$\,\pm$0.8} & 74.0{\fontsize{4}{5}\selectfont$\,\pm$6.7} & 46.9{\fontsize{4}{5}\selectfont$\,\pm$1.0} & 80.7{\fontsize{4}{5}\selectfont$\,\pm$4.8} & 73.3{\fontsize{4}{5}\selectfont$\,\pm$2.9} & 67.9{\fontsize{4}{5}\selectfont$\,\pm$1.0} & 47.2{\fontsize{4}{5}\selectfont$\,\pm$0.8} \\  %
\rowcolor{ourrow} RARA & \textbf{86.7}{\fontsize{4}{5}\selectfont$\,\pm$2.0} & \textbf{63.3}{\fontsize{4}{5}\selectfont$\,\pm$4.2} & \textbf{51.6}{\fontsize{4}{5}\selectfont$\,\pm$2.5} & \textbf{23.8}{\fontsize{4}{5}\selectfont$\,\pm$1.4} & 80.4{\fontsize{4}{5}\selectfont$\,\pm$3.3} & \textbf{64.2}{\fontsize{4}{5}\selectfont$\,\pm$2.7} & \textbf{85.3}{\fontsize{4}{5}\selectfont$\,\pm$0.7} & \textbf{82.0}{\fontsize{4}{5}\selectfont$\,\pm$2.4} & \textbf{76.0}{\fontsize{4}{5}\selectfont$\,\pm$1.0} & \textbf{58.3}{\fontsize{4}{5}\selectfont$\,\pm$1.2} \\  %
\bottomrule
\end{tabularx}

\vspace{0.6em}
\resizebox{\linewidth}{!}{%
\begin{tabular}{@{}l*{12}{c}@{}}
\toprule
\textbf{RoboCasa} & \multicolumn{2}{c}{\shortstack{Close Fridge\\Drawer}} & \multicolumn{2}{c}{\shortstack{Open\\Blender Lid}} & \multicolumn{2}{c}{\shortstack{Open\\Cabinet}} & \multicolumn{2}{c}{\shortstack{Turn On\\Toaster}} & \multicolumn{2}{c}{\shortstack{Pick-Place\\to Drawer}} & \multicolumn{2}{c}{\shortstack{Task\\average}} \\
\cmidrule(lr){2-3}\cmidrule(lr){4-5}\cmidrule(lr){6-7}\cmidrule(lr){8-9}\cmidrule(lr){10-11}\cmidrule(lr){12-13}
Method & ID & OOD & ID & OOD & ID & OOD & ID & OOD & ID & OOD & ID & OOD \\
\midrule
FT & 86.0{\fontsize{4}{5}\selectfont$\,\pm$8.7} & 62.7{\fontsize{4}{5}\selectfont$\,\pm$3.1} & \underline{76.0}{\fontsize{4}{5}\selectfont$\,\pm$3.5} & 44.7{\fontsize{4}{5}\selectfont$\,\pm$9.0} & \underline{72.0}{\fontsize{4}{5}\selectfont$\,\pm$2.0} & 44.0{\fontsize{4}{5}\selectfont$\,\pm$10.6} & \underline{72.0}{\fontsize{4}{5}\selectfont$\,\pm$5.3} & 46.0{\fontsize{4}{5}\selectfont$\,\pm$6.0} & 40.0{\fontsize{4}{5}\selectfont$\,\pm$2.0} & 24.0{\fontsize{4}{5}\selectfont$\,\pm$8.0} & \underline{69.2}{\fontsize{4}{5}\selectfont$\,\pm$2.5} & 44.3{\fontsize{4}{5}\selectfont$\,\pm$2.0} \\  %
Co-FT & \underline{90.7}{\fontsize{4}{5}\selectfont$\,\pm$3.1} & 59.3{\fontsize{4}{5}\selectfont$\,\pm$9.5} & 72.0{\fontsize{4}{5}\selectfont$\,\pm$5.3} & 48.0{\fontsize{4}{5}\selectfont$\,\pm$5.3} & 66.7{\fontsize{4}{5}\selectfont$\,\pm$9.0} & 47.3{\fontsize{4}{5}\selectfont$\,\pm$2.3} & 66.7{\fontsize{4}{5}\selectfont$\,\pm$11.0} & 42.0{\fontsize{4}{5}\selectfont$\,\pm$3.5} & 39.3{\fontsize{4}{5}\selectfont$\,\pm$1.2} & 22.0{\fontsize{4}{5}\selectfont$\,\pm$3.5} & 67.1{\fontsize{4}{5}\selectfont$\,\pm$0.5} & 43.7{\fontsize{4}{5}\selectfont$\,\pm$2.0} \\  %
STRAP & 84.0{\fontsize{4}{5}\selectfont$\,\pm$8.7} & 66.7{\fontsize{4}{5}\selectfont$\,\pm$3.1} & 73.3{\fontsize{4}{5}\selectfont$\,\pm$1.2} & 46.7{\fontsize{4}{5}\selectfont$\,\pm$3.1} & 68.7{\fontsize{4}{5}\selectfont$\,\pm$1.2} & \underline{52.0}{\fontsize{4}{5}\selectfont$\,\pm$5.3} & 66.7{\fontsize{4}{5}\selectfont$\,\pm$6.1} & \underline{47.3}{\fontsize{4}{5}\selectfont$\,\pm$7.0} & 38.0{\fontsize{4}{5}\selectfont$\,\pm$5.3} & \underline{28.0}{\fontsize{4}{5}\selectfont$\,\pm$7.2} & 66.1{\fontsize{4}{5}\selectfont$\,\pm$3.7} & \underline{48.1}{\fontsize{4}{5}\selectfont$\,\pm$3.2} \\  %
BlindVLA & 89.3{\fontsize{4}{5}\selectfont$\,\pm$1.2} & \underline{68.0}{\fontsize{4}{5}\selectfont$\,\pm$4.0} & 72.0{\fontsize{4}{5}\selectfont$\,\pm$12.0} & \underline{50.7}{\fontsize{4}{5}\selectfont$\,\pm$1.2} & 71.3{\fontsize{4}{5}\selectfont$\,\pm$6.1} & 47.3{\fontsize{4}{5}\selectfont$\,\pm$9.0} & 68.7{\fontsize{4}{5}\selectfont$\,\pm$16.2} & 44.7{\fontsize{4}{5}\selectfont$\,\pm$7.0} & \underline{41.3}{\fontsize{4}{5}\selectfont$\,\pm$2.3} & 23.3{\fontsize{4}{5}\selectfont$\,\pm$5.0} & 68.5{\fontsize{4}{5}\selectfont$\,\pm$5.4} & 46.8{\fontsize{4}{5}\selectfont$\,\pm$1.1} \\  %
\rowcolor{ourrow} RARA & \textbf{92.0}{\fontsize{4}{5}\selectfont$\,\pm$4.0} & \textbf{76.0}{\fontsize{4}{5}\selectfont$\,\pm$6.9} & \textbf{80.7}{\fontsize{4}{5}\selectfont$\,\pm$3.1} & \textbf{56.7}{\fontsize{4}{5}\selectfont$\,\pm$9.0} & \textbf{78.0}{\fontsize{4}{5}\selectfont$\,\pm$5.3} & \textbf{55.3}{\fontsize{4}{5}\selectfont$\,\pm$3.1} & \textbf{74.0}{\fontsize{4}{5}\selectfont$\,\pm$4.0} & \textbf{52.7}{\fontsize{4}{5}\selectfont$\,\pm$4.6} & \textbf{45.3}{\fontsize{4}{5}\selectfont$\,\pm$3.1} & \textbf{36.7}{\fontsize{4}{5}\selectfont$\,\pm$16.0} & \textbf{74.0}{\fontsize{4}{5}\selectfont$\,\pm$1.4} & \textbf{55.5}{\fontsize{4}{5}\selectfont$\,\pm$4.8} \\  %
\bottomrule
\end{tabular}
}
\vspace{-0.8em}
\end{table}

\begin{figure}[b]
    \centering
    \includegraphics[width=\linewidth]{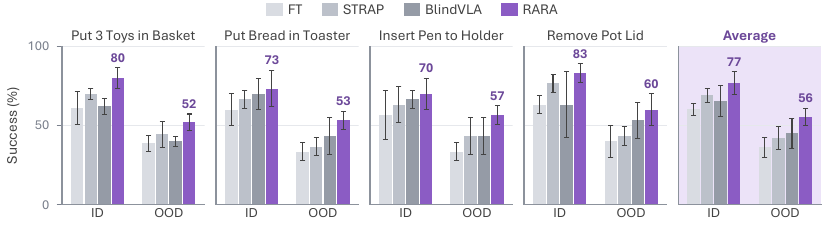}
    \vspace{-1.2em}
    \caption{Real-robot results (success rate, \%, mean\,$\pm$\,std over three runs) with GR00T~N1.7 pre-trained on DROID. RARA outperforms baselines in all target tasks, ID and OOD settings.}
    \label{fig:real_results}
    \vspace{-0.8em}
\end{figure}

Anchoring changes the representation of the retrieved pre-training frames as little as possible by design. RQ1 asks whether RARA also keeps the representation of target-task frames, on which the anchor is never applied, close to that of the pre-trained policy $P$, and whether it keeps the structure that $P$ learned in pre-training. We compare the anchored representation of each adapted policy with that of $P$ on Move Pill Bottle and Place Container in RoboTwin, whose objects do not appear in the pre-training data (Table~\ref{tab:rq1}; see Appx.~\ref{app:rq1_metrics} for definitions). The left part of Table~\ref{tab:rq1} is computed on held-out pre-training frames, and higher values mean that more of the pre-trained structure remains: \emph{effective rank}~\citep{garrido2023rankme} measures how many directions the representation uses, 
\emph{CKA}~\citep{kornblith2019similarity} measures its similarity to that of $P$,
and the \emph{position-probe $R^2$}~\citep{davari2022probing} measures how well object positions can be read from it. The right part reports \emph{feature distortion}~\citep{kumar2022finetuning} on frames from held-out episodes of the target task, in the ID and OOD conditions; it measures how far adaptation moves the representation from that of $P$, and lower values mean smaller changes.

\paragraph{RARA keeps the representation of target-task frames close to that of the pre-trained policy $P$.}
On both tasks and in both ID and OOD conditions, FT and Co-FT change the representation of target-task frames by $0.47$--$0.63$, whereas RARA changes it by only $0.23$--$0.33$, although the anchor is never applied to these frames. On pre-training frames, FT lowers the rank, CKA and probe $R^2$ on both tasks, whereas RARA stays close to $P$ (CKA $0.91$--$0.92$, $R^2$ $0.70$--$0.71$). Co-FT, which trains on the retrieved demonstrations with their actions, also keeps a high CKA ($0.93$--$0.94$), but it changes the representation of target-task frames the most, and its success is similar to or lower than that of FT (Table~\ref{tab:main}). Keeping the structure of the pre-training frames is therefore not sufficient; RARA also keeps the target-task representation close to that of $P$, while improving success on both tasks.

\paragraph{Stage~1 aligns the representation and the Stage-2 anchor keeps it close to $P$.}
Right after Stage~1, the representation of target-task frames is about as far from that of $P$ as after FT (distortion $0.46$--$0.56$, row ``RARA, after Stage~1''). Stage~1 does not keep the representation close to $P$; it aligns the target-task frames with the retrieved frames. Fine-tuning with the anchor in Stage~2 then reduces the distortion to $0.23$--$0.33$. RQ4 shows that both stages are needed for the full gain.

\subsection{RQ2: Does RARA Improve ID and OOD Success?}
\label{sec:exp_main}

This experiment tests the main claim of the paper: keeping the representation close to retrieved pre-training data should improve success under distribution shift, not only on the configuration seen in the demonstrations. We therefore report ID and OOD success separately (Table~\ref{tab:main} for simulation and Fig.~\ref{fig:real_results} for the real robot).

RARA achieves the highest task-average success in ID and in OOD in all three settings. Over FT, it gains $8.6$ (ID) and $11.0$ (OOD) points on RoboTwin, $4.8$ and $11.2$ points on RoboCasa, and $16.4$ and $19.2$ points on the real robot. In all three settings, the gain is larger in OOD than in ID, which is the goal of robust fine-tuning. The baselines, which reuse pre-training knowledge in other ways, improve less. Co-FT and STRAP, which train on pre-training demonstrations with their actions, gain at most $5.5$ points in OOD, and Co-FT does not consistently improve over FT in simulation. RETAIN, which merges weights, falls well below FT in simulation (see Appx.~\ref{app:retain} for a discussion), and BlindVLA, which aligns with a frozen visual encoder, gains at most $2.5$ points in simulation and $8.6$ points on the real robot. Since these baselines differ from RARA in more than one respect, we evaluate the contribution of each component of RARA separately in RQ4.

\subsection{RQ3: Does the Choice of Anchored Layer Matter?}
\label{sec:exp_layers}

\begin{wrapfigure}{r}{0.60\textwidth}
    \vspace{-1.2em}
    \centering
    \includegraphics[width=\linewidth]{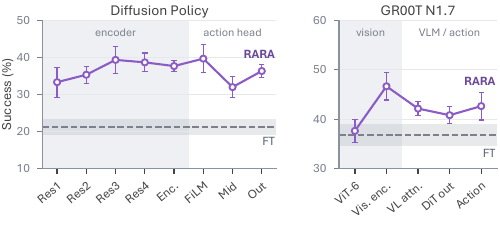}
    \vspace{-1.6em}
    \caption{Choice of anchored layer on Move Pill Bottle: mean of ID and OOD success. Left: Diffusion Policy. Right: GR00T~N1.7, where every arm trains the vision encoder and the action head with the LLM frozen, as FT does (Appx.~\ref{app:groot_layers}). Bars and band: std over the last three checkpoints.}
    \label{fig:anchor_location}
    \vspace{-1.2em}
\end{wrapfigure}
RARA aligns and anchors the representation at a single anchored layer, and the method does not fix which layer this is (\S\ref{sec:alignment}).
This experiment asks whether the choice matters. If only a few layers worked, a user would first have to search for the right layer before applying RARA.

We place the anchored layer at eight locations in the Diffusion Policy on Move Pill Bottle:
the four ResNet stages, the encoder output (our default), and three places in the action head (the FiLM conditioning, the middle block and the output). 

Every location improves over FT (Fig.~\ref{fig:anchor_location}, left): success rises from $21.2\%$ to between $32.0\%$ and $39.7\%$, a gain of $11$ to $18$ points. Some locations work better than others, but none of them comes close to FT. On this task, the gain therefore does not depend on one particular layer. GR00T~N1.7 shows a similar trend (Fig.~\ref{fig:anchor_location}, right): four of the five locations improve over FT by $4$ to $10$ points, and only an early layer of the vision encoder stays at the level of FT (Appx.~\ref{app:groot_layers}).

\subsection{RQ4: Which Components and Which Data Produce the Gain?}
\label{sec:exp_ablation}

We run an ablation study on RoboTwin, averaged over its four tasks, and on Open Blender Lid in RoboCasa (Table~\ref{tab:ablation_results}). We compare RARA with five variants. \emph{Stage~1 only} fine-tunes without the anchor after Stage~1, and \emph{Stage~2 only} skips Stage~1. \emph{Co-FT in Stage~2} adds behavior cloning on the retrieved demonstrations to Stage~2. \emph{Random pre-training data} replaces the retrieved demonstrations with a random subset of the pre-training data of the same size. \emph{Random Stage-1 pairs} pairs target and retrieved frames at random in Stage~1, instead of by action-chunk similarity.

\begin{wraptable}{r}{0.52\textwidth}
    \vspace{-1.4em}
    \centering
    \captionof{table}{Ablations of RARA (success rate, \%, mean\,$\pm$\,std). RoboTwin: average over its four tasks. Bold: best; underline: second best.}
\label{tab:ablation_results}
\resizebox{\linewidth}{!}{\scriptsize\setlength{\tabcolsep}{3pt}%
\begin{tabular}{@{}lcccc@{}}
\toprule
 & \multicolumn{2}{c}{RoboTwin (avg.)} & \multicolumn{2}{c}{Open Blender Lid} \\
\cmidrule(lr){2-3}\cmidrule(lr){4-5}
Method & ID & OOD & ID & OOD \\
\midrule
FT & 67.4{\fontsize{4.2}{5}\selectfont$\,\pm$1.4} & 47.3{\fontsize{4.2}{5}\selectfont$\,\pm$2.4} & 76.0{\fontsize{4.2}{5}\selectfont$\,\pm$3.5} & 44.7{\fontsize{4.2}{5}\selectfont$\,\pm$9.0} \\
\rowcolor{ourrow} RARA & \textbf{76.0}{\fontsize{4.2}{5}\selectfont$\,\pm$1.0} & \textbf{58.3}{\fontsize{4.2}{5}\selectfont$\,\pm$1.2} & \textbf{80.7}{\fontsize{4.2}{5}\selectfont$\,\pm$3.1} & \textbf{56.7}{\fontsize{4.2}{5}\selectfont$\,\pm$9.0} \\
\quad Random pre-training data & 50.3{\fontsize{4.2}{5}\selectfont$\,\pm$1.7} & 34.9{\fontsize{4.2}{5}\selectfont$\,\pm$1.7} & 67.7{\fontsize{4.2}{5}\selectfont$\,\pm$1.3} & 44.2{\fontsize{4.2}{5}\selectfont$\,\pm$1.2} \\
\quad Random Stage-1 pairs & 65.3{\fontsize{4.2}{5}\selectfont$\,\pm$3.1} & 40.6{\fontsize{4.2}{5}\selectfont$\,\pm$1.2} & 76.7{\fontsize{4.2}{5}\selectfont$\,\pm$2.3} & 45.5{\fontsize{4.2}{5}\selectfont$\,\pm$4.2} \\
\quad Co-FT in Stage~2 & 65.8{\fontsize{4.2}{5}\selectfont$\,\pm$3.3} & \underline{48.1}{\fontsize{4.2}{5}\selectfont$\,\pm$2.9} & 72.9{\fontsize{4.2}{5}\selectfont$\,\pm$5.3} & \underline{53.6}{\fontsize{4.2}{5}\selectfont$\,\pm$3.7} \\
\quad Stage~1 only & 66.6{\fontsize{4.2}{5}\selectfont$\,\pm$2.2} & 43.2{\fontsize{4.2}{5}\selectfont$\,\pm$1.3} & \underline{77.3}{\fontsize{4.2}{5}\selectfont$\,\pm$4.8} & 52.2{\fontsize{4.2}{5}\selectfont$\,\pm$2.3} \\
\quad Stage~2 only & \underline{73.0}{\fontsize{4.2}{5}\selectfont$\,\pm$3.2} & 46.2{\fontsize{4.2}{5}\selectfont$\,\pm$4.1} & 76.0{\fontsize{4.2}{5}\selectfont$\,\pm$6.0} & 48.0{\fontsize{4.2}{5}\selectfont$\,\pm$7.2} \\
\bottomrule
\end{tabular}}

    \vspace{-1em}
\end{wraptable}
Both stages are needed. On RoboTwin, Stage~1 only is no better than FT, and Stage~2 only recovers most of the ID gain ($73.0\%$ vs.\ $76.0\%$) but none of the OOD gain ($46.2\%$, vs.\ $47.3\%$ for FT and $58.3\%$ for RARA). On Open Blender Lid, each stage alone recovers part of the OOD gain, and only the full method reaches $56.7\%$. 
A likely explanation is that each stage needs the other. Without the anchor (Stage~1 only), fine-tuning is free to move the representation again, so the alignment from Stage~1 is not preserved and the policy behaves like FT. Without Stage~1 (Stage~2 only), the anchor preserves the pre-trained representation of the retrieved demonstrations, but the target frames are not aligned with it, so the policy cannot reuse this robust representation in OOD conditions.
Adding behavior cloning on the retrieved demonstrations (Co-FT in Stage~2) lowers success in all four columns, so the retrieved demonstrations are most useful as an anchor for the representation.

The choice of pre-training data and how its frames are paired also matter. With random pre-training data, RARA performs worse than FT on both RoboTwin and Open Blender Lid, by up to $17.1$ points on RoboTwin. With random Stage-1 pairs, RARA falls below FT on RoboTwin ($65.3\%$/$40.6\%$ vs.\ $67.4\%$/$47.3\%$) and is only on par with FT on Open Blender Lid, so the action-aware pairing is needed. Appx.~\ref{sec:exp_skillmatch} further shows that the anchor helps when the pre-training data contain the same skill as the target task, and can hurt when the skill differs.

\section{Conclusion}
\label{sec:conclusion}

We presented RARA, a two-stage method for robust policy fine-tuning. RARA retrieves pre-training demonstrations of the same type of task as the target, first aligns the target representation to their pre-trained representation, and then fine-tunes the policy while anchoring the representation of these demonstrations. Across 13 tasks in simulation and on a real robot, RARA improves the average success of fine-tuning by $9.5$ points in ID and $13.6$ points in OOD. Our analysis shows that it also keeps the representation of target-task frames closer to that of the pre-trained policy, although the anchor is applied only to the retrieved demonstrations. While being effective, RARA relies on pre-training data that contain the same type of task as the target, and our experiments focus on visual changes with the same robot and action space. Future work includes retrieving useful references when such data are scarce, and extending RARA to new embodiments and action spaces.

\subsubsection*{Acknowledgments}
This work was partly supported by Delta Electronics Inc. and Toyota Research Institute.
This work used the Delta system at the National Center for Supercomputing Applications [award OAC 2005572] through allocation CIS261341 from the Advanced Cyberinfrastructure Coordination Ecosystem: Services \& Support (ACCESS) program~\citep{boerner2023access}, which is supported by National Science Foundation grants \#2138259, \#2138286, \#2138307, \#2137603, and \#2138296.

\bibliographystyle{iclr2027_conference}
\bibliography{ref}  %

\newpage
\appendix

\section{Analysis of Visual Feature Shift}
\label{app:visual_shift_analysis}
\label{sec:visual_shift_analysis}

\begin{figure}[h]
    \centering
    \includegraphics[width=0.5\linewidth]{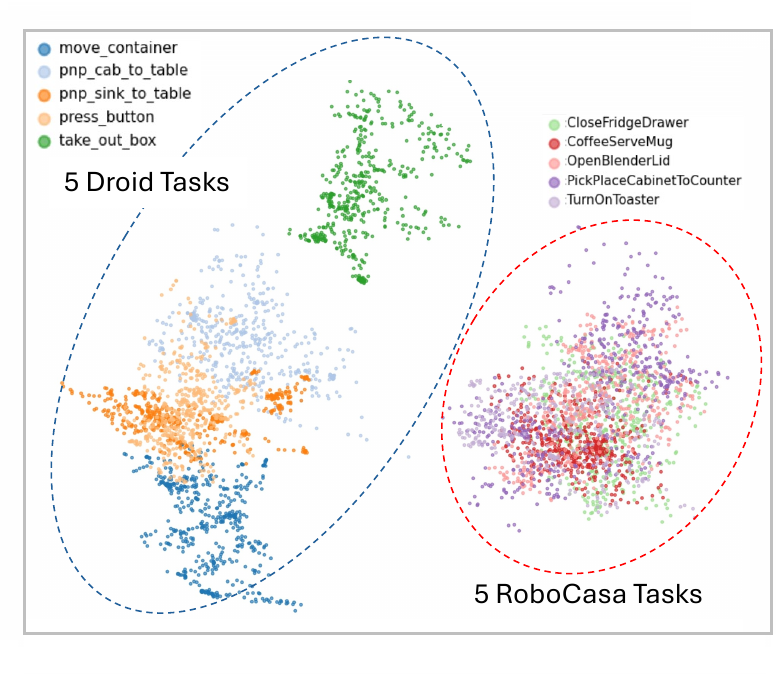}
    \caption{PCA visualization of visual features of ID and OOD tasks.}
    \label{fig:visual_feature_shift}
\end{figure}

We discuss visual feature alignment for VLA adaptation. A pre-trained VLA visual encoder can separate diverse source tasks, but downstream images from a new simulator or real setup can look different. Even when the action space and language instructions are similar, such image shifts can leave target features poorly organized for action prediction.

Fig.~\ref{fig:visual_feature_shift} illustrates this issue. We extract features from a DROID-pre-trained $\pi_0$ policy and visualize them with PCA, using five DROID tasks as ID source tasks and five RoboCasa tasks as OOD downstream tasks~\citep{robocasa365}; details are in \S\ref{app:pca_details}. DROID features are well separated, while unseen RoboCasa features collapse into a compact region despite representing different manipulation tasks. Thus, the DROID-pre-trained visual encoder is not sufficiently task-discriminative in the unseen environment.

Such feature collapse is problematic for policy adaptation. If the visual encoder maps different target states or tasks into a collapsed region, the action model receives weak state information and can easily overfit the few available demonstrations. This motivates visual feature alignment: before learning target actions, we first move target-domain visual features toward a task-relevant region of the pre-trained feature space, making the new observations easier for the pre-trained policy to interpret and providing a better starting point for adaptation.

\subsection{PCA Visualization Details}
\label{app:pca_details}

For the $\pi_0$ feature visualizations, we extract image features from
the SigLIP visual encoder before any pooling operation. For each image
and each camera view, the encoder outputs 196 patch tokens with feature
dimension 1152. We average over the 196 patch tokens to obtain one
1152-dimensional feature vector per camera. The three camera-specific
vectors are then concatenated, giving a 3456-dimensional visual feature
for each multi-view observation. Stacking features over all selected
observations yields a matrix in $\mathbb{R}^{N\times3456}$. We fit PCA
to this feature matrix using SVD and project each observation onto the
top two principal components for visualization. When comparing multiple
data sources or adaptation checkpoints, we fit PCA once on the combined
feature set and use the same projection for all points.

\section{Retrieval and Pair Construction Details}
\label{app:source_curation_details}

This appendix details how RARA retrieves pre-training demonstrations and pairs their frames with target frames (\S\ref{sec:retrieval}). The same procedure is used for DROID and for RoboTwin.

\paragraph{Retrieval by instruction.}
Each demonstration comes with a task instruction, and a target task usually has several instructions; on RoboTwin, they are generated from templates. Let $e(g)$ be the embedding of an instruction $g$ from a frozen CLIP text encoder~\citep{radford2021learning}. For a target task with instruction set $\mathcal{G}_{\mathrm{tgt}}$, we average the normalized instruction embeddings,
\begin{equation}
    \bar{e}_{\mathrm{tgt}}
    = \frac{1}{|\mathcal{G}_{\mathrm{tgt}}|}
    \sum_{g \in \mathcal{G}_{\mathrm{tgt}}}
    \frac{e(g)}{\|e(g)\|_2}.
\end{equation}
We score each pre-training demonstration by the cosine similarity between the embedding of its instruction and $\bar{e}_{\mathrm{tgt}}$, and keep the top $K=100$ demonstrations as $D_{\mathrm{ret}}$. Filtering by instruction separates tasks with different intent, such as opening, closing, placing, wiping and insertion, even when their action chunks contain similar end-effector displacements.

\paragraph{Action chunks.}
We represent each demonstration as $\tau=\left(g,\{o_t,q_t,a_t\}_{t=1}^{T}\right)$, where $g$ is the instruction, $o_t$ the image observation, $q_t$ the proprioceptive state, and $a_t$ the action in the action space of the policy. Each demonstration is split into overlapping action chunks
\begin{equation}
    A_t = [a_t, a_{t+1}, \ldots, a_{t+H-1}],
    \qquad t \in \{1, 1+s, 1+2s, \ldots\},
\end{equation}
with horizon $H=16$ and stride $s$, and each chunk is paired with the observation at its first step. Continuous action dimensions are normalized with statistics of the pre-training data, and the gripper dimension is scaled separately so that binary open/close changes do not dominate the translational and rotational dimensions. Target chunks are normalized in the same way.

\paragraph{DTW distance.}
For two action chunks $A$ and $B$, we use the dynamic-time-warping distance
\begin{equation}
    \delta(A,B)
    = \min_{\omega \in \Omega(A,B)}
    \frac{1}{|\omega|}
    \sum_{(u,v) \in \omega}
    \left\| A_u - B_v \right\|_2,
\end{equation}
where $\Omega(A,B)$ is the set of monotone temporal alignments between the two chunks. DTW matches similar actions that are executed at different speeds.

\paragraph{Positive and negative pairs.}
For each target chunk $A_i^{\mathrm{tgt}}$, we compute $\delta$ to every chunk of the demonstrations in $D_{\mathrm{ret}}$. The $m_{+}$ closest chunks are the positives $\mathcal{P}(i)$; they can come from any retrieved demonstration. The negatives $\mathcal{N}(i)$ are sampled from the retrieved chunks whose distance to $A_i^{\mathrm{tgt}}$ is larger than the median of these distances. Each chunk is represented by the observation at its first step. Because pairs are drawn only from $D_{\mathrm{ret}}$, a target frame is not paired with a frame of an unrelated task that happens to have similar actions, such as a pushing or placing segment for a lid-opening target task.

\begin{figure}[h]
    \centering
    \includegraphics[width=\linewidth]{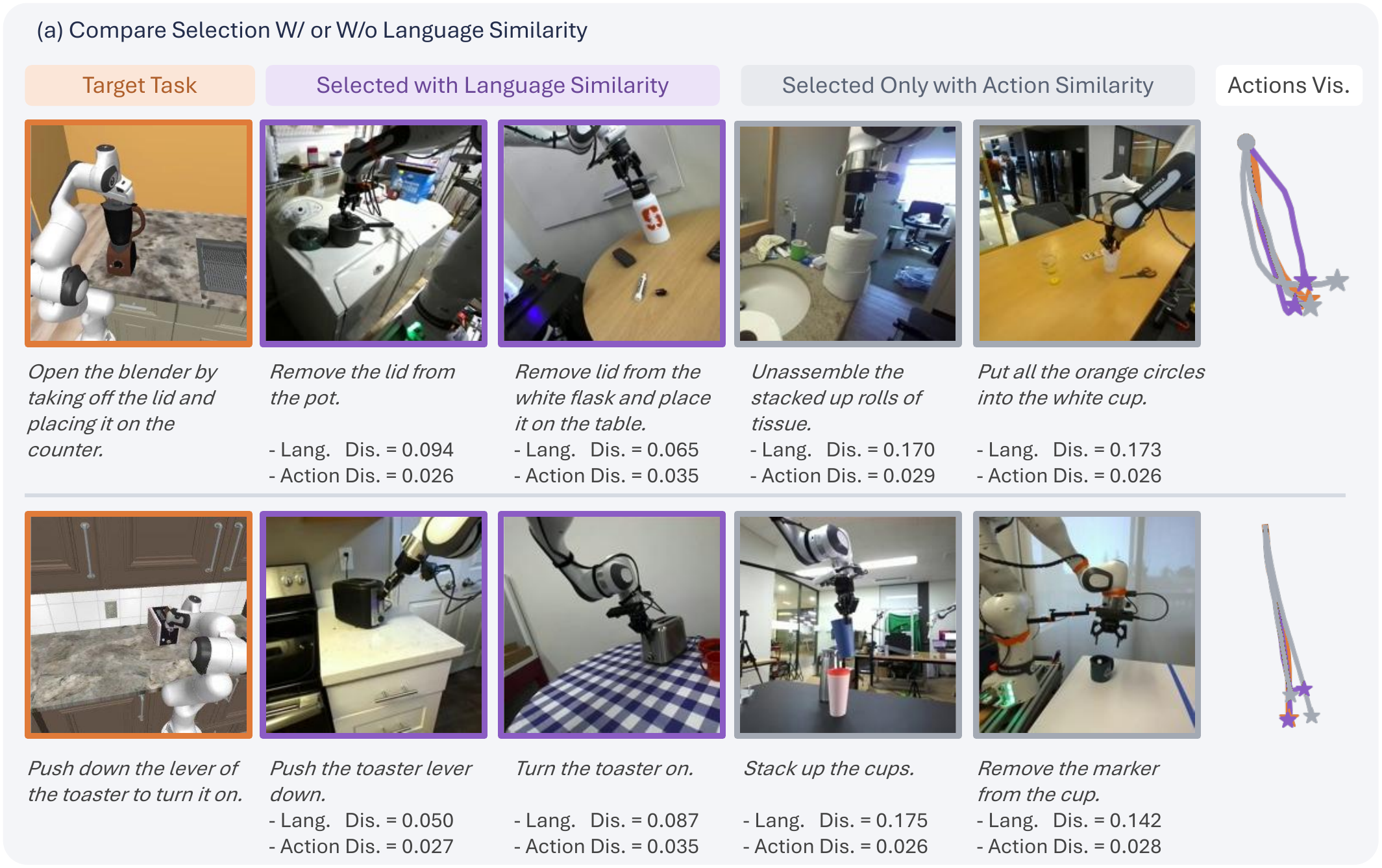}
    \caption{Additional examples of retrieval for RARA. Without the instruction filter, selection by action similarity can return pre-training demonstrations with similar action chunks but a different task. Filtering by the task instruction returns demonstrations that better match the target task.}
    \label{fig:app_task_conditioned_selection_examples}
\end{figure}

Figure~\ref{fig:app_task_conditioned_selection_examples} shows additional examples of retrieval with and without the instruction filter.

\section{Implementation Details}
\label{app:implementation_details}

\paragraph{Anchored representation.}
Our target tasks differ from the pre-training data mainly in visual appearance, so we anchor the representation computed from the visual input. For the Diffusion Policy, this is the pooled output of the ResNet-18 encoders of all cameras, concatenated into a 1024-dimensional feature. For GR00T, it is the output tokens of the vision tower. Stage~1 trains the layers that produce this representation, together with the projector (Appx.~\ref{app:projector}), for 3\,000 steps, and keeps the rest of the policy fixed.

\paragraph{Anchor distance.}
The distance $d$ in the anchor loss (Eq.~\ref{eq:anchor_loss}) is the mean squared error for the Diffusion Policy. For GR00T, we divide the squared error by the batch variance of the reference representation $\bar h^{\ell}$, so that the same weight can be used for representations of different scales.

\paragraph{Retrieval and training budget.}
In all settings, we retrieve $K=100$ demonstrations and use action chunks of $H=16$ steps (Appx.~\ref{app:source_curation_details}). All methods in a setting use the same optimizer settings and training budget. Appx.~\ref{app:robotwin} lists the full settings of the Diffusion Policy on RoboTwin.

\section{Evaluation Environments}
\label{app:evaluation_envs}

This appendix describes the RoboCasa and real-robot settings (\S\ref{app:robocasa_real}), the RoboTwin setting (\S\ref{app:robotwin}), and the representation metrics used in RQ1 (\S\ref{app:rq1_metrics}).

\subsection{RoboCasa and Real Robot}
\label{app:robocasa_real}

\begin{figure}[h]
    \centering
    \includegraphics[width=\linewidth]{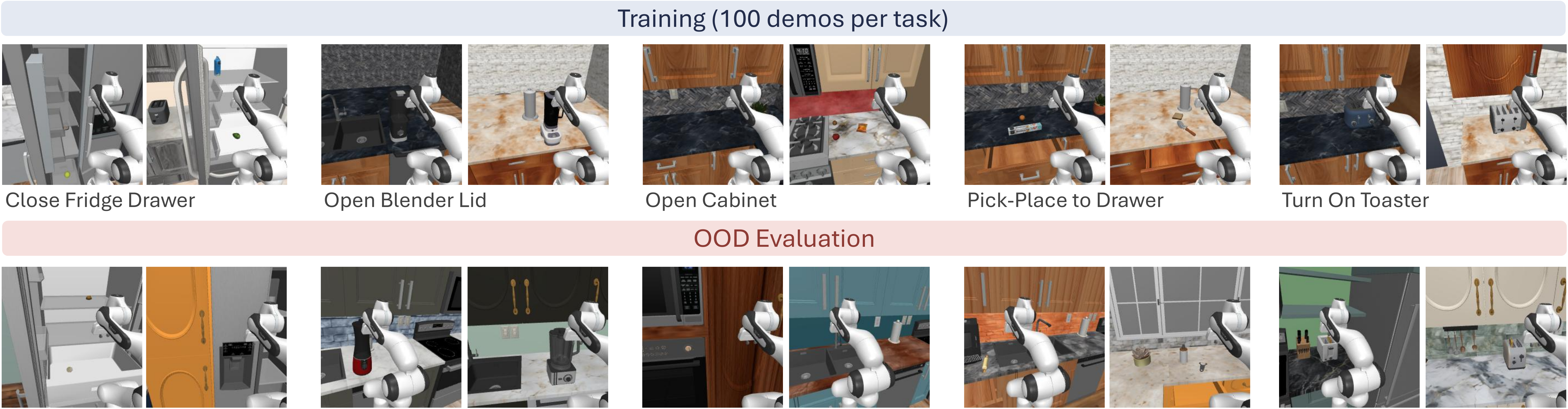}
    \caption{RoboCasa training and OOD evaluation environments for all simulated robot tasks. OOD settings include variations in camera view, background, distractors, objects, clutter, and layouts.}
    \label{fig:app_robocasa_id_ood_vis}
\end{figure}

\begin{figure}[h]
    \centering
    \includegraphics[width=\linewidth]{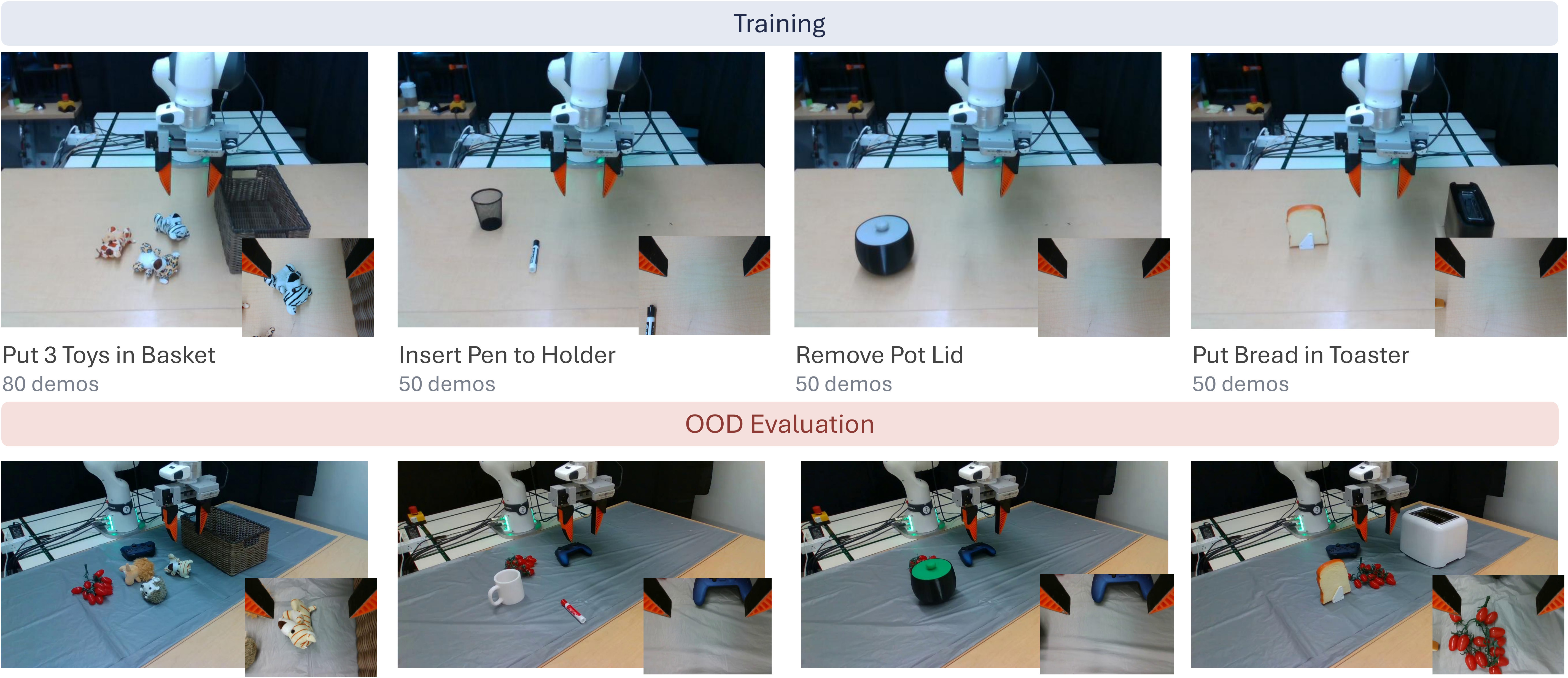}
    \caption{Real-world Franka training and OOD evaluation environments for all robot tasks. OOD settings include variations in camera view, background, distractors, and objects.}
    \label{fig:app_real_world_id_ood_vis}
\end{figure}

Figures~\ref{fig:app_robocasa_id_ood_vis} and~\ref{fig:app_real_world_id_ood_vis} show the training and OOD evaluation environments of all RoboCasa and real-robot tasks.

\paragraph{Evaluation protocol.}
In RoboCasa, each policy is evaluated for 50 episodes per task, and each experiment is repeated with three training seeds. For GR00T-based policies, the model predicts an action chunk of 16 steps and executes all 16 steps before replanning. In the real-world Franka experiments, each trained policy is evaluated for 10 trials per task, and this evaluation is repeated three times. For Put 3 Toys in Basket, each trial is scored by the number of toys placed in the basket, so the success rate is the fraction of the three toys placed. To improve fairness across real-world comparisons, we record the randomization setting for each trial, including scene layout, object placement, distractors, and other visual variations. We then evaluate the next model under the same recorded setting whenever possible before switching to a new randomized setting. For real-world GR00T-based policies, the model predicts a 16-step action chunk but executes only the first 8 steps before replanning.

\FloatBarrier
\subsection{RoboTwin}
\label{app:robotwin}

RoboTwin~2.0~\citep{chen2025robotwin2} makes it easy to collect demonstrations, and its randomization of object instances, backgrounds, clutter and object positions gives enough diversity to build both a broad pre-training set and narrower target tasks in the same environment. We use the right arm of the Aloha-AgileX embodiment with the head and wrist RGB cameras.

\paragraph{Pre-training and target data.}
We collect about 2{,}000 pre-training demonstrations in total, in two groups, each paired with two target tasks.

The first group contains pick-and-place demonstrations (\texttt{place\_a2b\_right}), in which the right arm picks up an object and places it next to a second object. The first object is drawn from 11 categories (playing cards, mouse, bell, stapler, soap, toy car, coffee box, phone, Rubik's cube, bread and tea box), and the second from six (Rubik's cube, wooden block, phone, bread, bell and soap). This group serves Place Container and Move Pill Bottle. These targets use the same skill, but their objects do not appear in the pre-training data: Move Pill Bottle uses a pill bottle and a blue pad, and Place Container uses a plate together with a bowl or a cup. The novelty of these two targets is therefore at the object level.

The second group contains four skills: pressing a bell (\texttt{click\_bell}), shaking a bottle vertically (\texttt{shake\_bottle}), opening a laptop (\texttt{open\_laptop}) and rotating a QR code (\texttt{rotate\_qrcode}). This group serves Click Alarm Clock and Shake Bottle. Click Alarm Clock uses the same pressing skill as \texttt{click\_bell} on a new object, and Shake Bottle shakes the same kind of object horizontally instead of vertically; neither target task appears in the pre-training data. Opening a laptop and rotating a QR code are unrelated to both targets, and retrieval is meant to exclude them.

\begin{figure}[h]
    \centering
    \includegraphics[width=\linewidth]{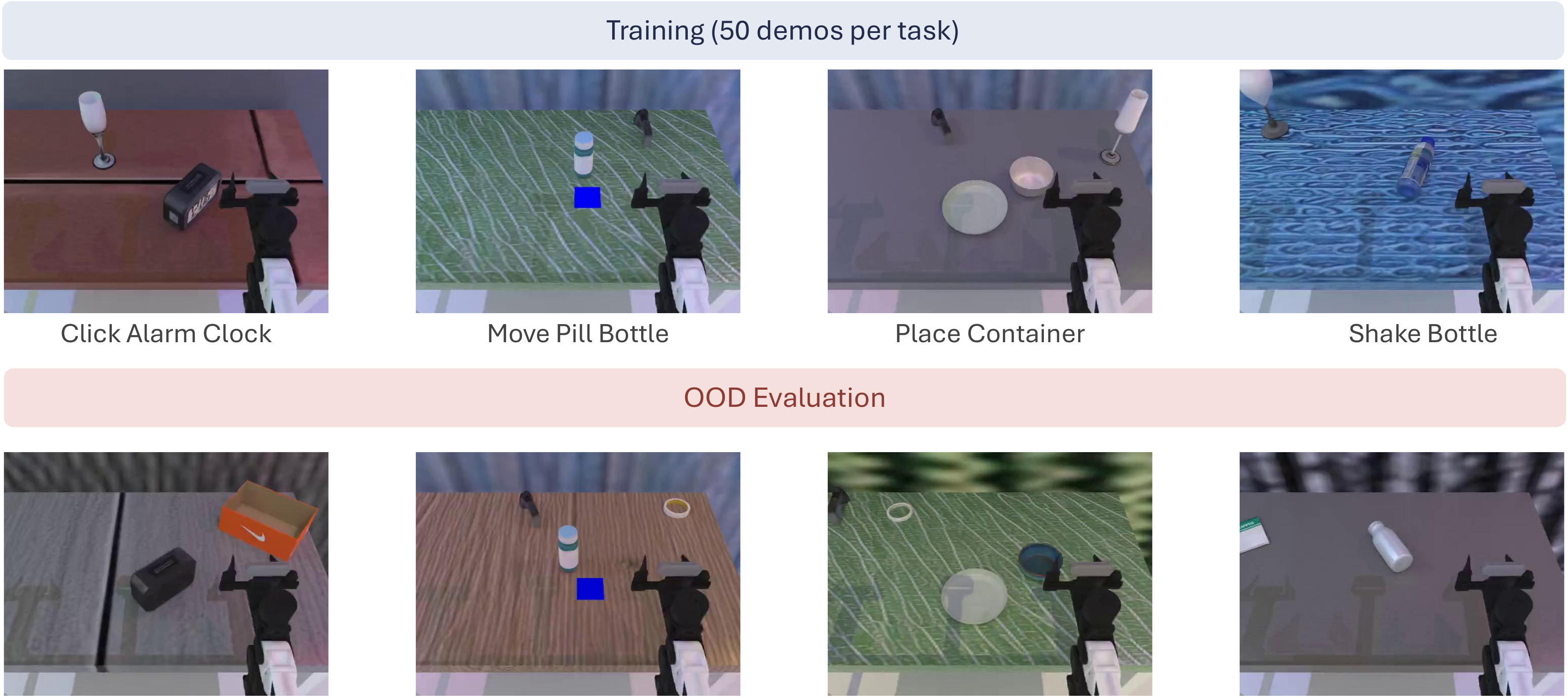}
    \caption{RoboTwin training and OOD evaluation environments for the four target tasks. OOD settings widen the object layout and vary the background and distractors.}
    \label{fig:app_robotwin_id_ood_vis}
\end{figure}

The pre-training demonstrations are collected with wide randomization: 100 table textures and backgrounds, cluttered distractors, and object positions over the whole reachable table. Each target task has 50 demonstrations, collected with a narrow layout, 10 backgrounds and light clutter. ID evaluation uses the narrow layout of the demonstrations; OOD evaluation widens the object layout, the backgrounds and the distractors. Figure~\ref{fig:app_robotwin_id_ood_vis} shows the training and OOD evaluation environments of the four target tasks.

\paragraph{Policy and training.}
\label{app:robotwin_hparams}
We use the Diffusion Policy~\citep{chi2023diffusion} implementation in XPolicyLab~\citep{xpolicylab2026}: a ResNet-18 encoder per camera with group normalization and spatial soft-argmax pooling (1024-dimensional visual feature over both cameras), a 1-D UNet denoiser, and 16-step action chunks. It is pre-trained for 120 epochs on the pre-training set. Every adaptation method then trains for 300 epochs on the 50 target demonstrations with a batch of 128. Co-FT fills each batch with 64 target chunks and 64 chunks of the retrieved pre-training demonstrations $D_{\mathrm{ret}}$, and applies the behavior-cloning loss to both. In Stage~2 of RARA, each batch holds 64 target chunks and 64 pre-training frames; the behavior-cloning loss is applied to the target chunks and $\mathcal{L}_{\mathrm{anchor}}$ (mean squared error, $\lambda_2=1$) to the pre-training frames, without their actions. The \emph{frozen} reference in RQ1 is FT with the layers that produce the anchored representation fixed. Table~\ref{tab:robotwin_hparams} lists the remaining settings.

\paragraph{Anchored layer.}
By default, the anchored representation $h^{\ell}$ is the pooled output of the two ResNet-18 encoders, concatenated into a 1024-dimensional feature. Stage~1 updates these encoders and the projector. Appx.~\ref{app:layer_choice} describes the other locations used in RQ3.

\paragraph{Stage~1.}
\label{app:robotwin_stage1}
Retrieval and pair construction follow \S\ref{sec:retrieval} and Appx.~\ref{app:source_curation_details}: the top $K=100$ pre-training demonstrations are retrieved by instruction similarity, and the positives and negatives are defined by the DTW distance between action chunks, with $m_{+}=4$. Stage~1 converges on both targets: the contrastive loss falls from $2.73$ to $0.22$ on Place Container ($92\%$) and from $2.62$ to $0.87$ on Move Pill Bottle ($67\%$).

\paragraph{Baselines.}
\label{app:robotwin_baselines}
All baselines start from the same pre-trained policy and use the same 50 target demonstrations and 300-epoch budget as FT. For weight-space preservation (RETAIN;~\citealp{yadav2025robust}), we interpolate the pre-trained and fine-tuned weights as $\theta(\alpha)=(1-\alpha)\theta_{\mathrm{pre}}+\alpha\theta_{\mathrm{ft}}$ with $\alpha\in\{0.25,0.5,0.75\}$. For same-image teacher alignment (BlindVLA;~\citealp{kachaev_dont_2025}), we add a feature-distillation term toward frozen DINOv2 features of each target image, with weight $\lambda\in\{1,0.1,0.01\}$. For retrieval and co-training (STRAP;~\citealp{memmel2024strap}), we follow the original method: the image observations are encoded with DINOv2, target sub-trajectories are matched to pre-training sub-trajectories by subsequence DTW over these visual features, and we keep the top 50 matches and train with the Co-FT loss and the same 1:1 batch mixture.

\paragraph{Evaluation.}
Rollouts use the benchmark's success check with a 400-step horizon. Each result uses the last three checkpoints (epochs 250, 275 and 300), each evaluated on 50 rollouts, so that checkpoint-to-checkpoint noise is averaged rather than selected. Each experiment is repeated with three training seeds, and we report the mean and standard deviation over the seeds.

\begin{table}[h]
\centering
\caption{Diffusion Policy settings on RoboTwin.}
\label{tab:robotwin_hparams}
\scriptsize
\begin{tabularx}{\linewidth}{@{}lX@{}}
\toprule
Anchored representation $h^{\ell}$ (default) & pooled encoder output of both cameras, concatenated (1024-d) \\
Other locations (RQ3) & ResNet stages 1--4, FiLM conditioning, UNet middle block, action output \\
\midrule
\multicolumn{2}{@{}l}{\emph{Stage~1 (alignment)}} \\
Trainable & encoder (both cameras) and projector \\
Projector $g_{\psi}$ & MLP $1024\!\to\!512\!\to\!256$, GELU, $\ell_2$-normalized output \\
Positives / negatives & $m_{+}=4$ nearest by action-chunk DTW; 64 negatives per step, sampled beyond the median distance \\
Temperature, $\lambda_1$ & $\tau=0.07$, $\lambda_1=1$ \\
Optimizer & AdamW, lr $10^{-5}$ (encoder) / $10^{-4}$ (projector), weight decay 0.01, 200 warm-up steps then cosine, gradient clip 1.0 \\
Steps, batch & 3\,000 steps; 32 target frames and 32 pre-training frames for $\mathcal{L}_{\mathrm{anchor}}$ per step \\
\midrule
\multicolumn{2}{@{}l}{\emph{Stage~2 (task training with the anchor)}} \\
Trainable & whole policy (encoder and UNet) \\
Batch & 128; target chunks and pre-training frames 1:1 \\
Distance $d$, $\lambda_2$ & mean squared error, $\lambda_2=1$ \\
Optimizer & AdamW, lr $10^{-4}$, cosine schedule, 300 epochs \\
Evaluation & epochs 250, 275, 300; 50 rollouts each per split \\
\bottomrule
\end{tabularx}
\end{table}

\subsection{RQ1 Metrics}
\label{app:rq1_metrics}

All metrics use the anchored representation, the pooled outputs of the two ResNet-18 encoders concatenated into a 1024-dimensional feature, on frames the policy was not trained on: 100 held-out target episodes in the ID region, 100 in the OOD region, and pre-grasp frames of held-out pre-training episodes. $f_{\theta}$ denotes the adapted encoder and $f_{P}$ the pre-trained one.

\emph{Feature distortion}~\citep{kumar2022finetuning} is $\mathbb{E}_{o}\,\|f_{\theta}(o)-f_{P}(o)\|_2/\|f_{P}(o)\|_2$ over the frames of a set; 0 means the features did not move. We divide by the norm of the pre-trained feature so that a change of scale does not dominate.

\emph{Effective rank}~\citep{garrido2023rankme} is $\exp(-\sum_i p_i\log p_i)$ with $p_i=\sigma_i/\sum_j\sigma_j$, where $\sigma_i$ are the singular values of the centered feature matrix of a set.

\emph{Linear CKA}~\citep{kornblith2019similarity} is $\|X^{\top}Y\|_F^2/(\|X^{\top}X\|_F\,\|Y^{\top}Y\|_F)$ for the centered feature matrices $X$ of the adapted and $Y$ of the pre-trained encoder on the same frames; 1 means the two representations agree up to rotation and scale. Because CKA can change without a change in function~\citep{davari2023reliability}, we do not rely on it alone.

\emph{Position-probe $R^2$}~\citep{davari2022probing} is the score of a ridge regression from the features to the object position at the grasp of the same episode (32 PCA components, $\alpha=10$, five-fold GroupKFold by episode). Frames after the grasp are excluded, because the object then moves with the hand. A drop relative to $P$ indicates forgetting.

\section{Choice of Anchored Layer}
\label{app:layer_choice}
\label{app:groot_layers}

This section gives the setup of the layer study in RQ3 (\S\ref{sec:exp_layers}, Fig.~\ref{fig:anchor_location}). Both policies are evaluated on Move Pill Bottle in RoboTwin, and each arm changes only the anchored layer.

\paragraph{Diffusion Policy.}
Besides the default location, the pooled output of the two ResNet-18 encoders, we anchor seven other locations: the outputs of the four ResNet stages (spatially pooled), the FiLM conditioning vector of the UNet, the output of the UNet middle block, and the predicted action. All arms use $\lambda_1=\lambda_2=1$. When the anchored layer is inside the action network, the diffusion timestep and the noisy action input are shared between the trainable and the reference policy, so the two differ only in their weights. All other settings follow Appx.~\ref{app:robotwin}.

\paragraph{GR00T.}
We use GR00T~N1.7 trained on the RoboTwin pre-training data. As in \S\ref{sec:method}, Stage~1 trains only the layers that produce the anchored representation. \textbf{In Stage~2, all anchor arms train the vision encoder and the action head with the LLM frozen, the same trainable parameters as the FT reference.} Any difference from FT therefore comes from where the anchor is applied, not from which parameters are fine-tuned. The LLM is frozen in Stage~2 because the official GR00T fine-tuning recipe keeps it frozen. To study anchored layers inside the LLM, such as the output of the vision-language self-attention, Stage~1 also trains the LLM layers that produce this representation. In our default setting, which anchors the output of the visual encoder, RARA improves over FT whether the LLM is frozen or trained. We anchor five locations: the output of layer 6 of the 24-layer ViT, the output of the visual encoder (our default), the output of the vision-language self-attention, the output of the DiT action head, and the predicted action. All arms use $\lambda_1=\lambda_2=0.1$.

\paragraph{Evaluation.}
For both policies, each point in Fig.~\ref{fig:anchor_location} averages the last three checkpoints, and the error bars show the standard deviation over these checkpoints. Each GR00T checkpoint is evaluated on 100 ID and 100 OOD rollouts.

\section{Projector Analysis}
\label{app:projector}

Stage~1 computes the contrastive loss in a projected space. A shared two-layer MLP projector ($1024\!\to\!512\!\to\!256$, GELU, $\ell_2$-normalized output) maps the anchored representation of both the target frames and the retrieved frames to the contrastive embedding, following common practice in contrastive learning~\citep{chen2020simple,bardes2022vicreg}. The projector is trained together with the policy (learning rate $10^{-4}$, compared with $10^{-5}$ for the policy layers) and is discarded after Stage~1, so Stage~2 starts from the updated policy without it. We compare this default with two alternatives on RoboTwin, using the policy pre-trained with one seed: no projector, where the loss is applied directly to the normalized representation, and a randomly initialized frozen projector, which \citet{kachaev_dont_2025} use when aligning with a frozen teacher. Each success rate is computed over 150 rollouts.

Removing the projector does not change the result of Stage~1 followed by fine-tuning without the anchor: $86.7\%$/$60.7\%$ vs.\ $87.3\%$/$60.0\%$ (ID/OOD) on Place Container, and $30.7\%$ vs.\ $32.0\%$ (ID) on Move Pill Bottle. A frozen projector changes Stage~1 itself. The representation aligns more strongly: on held-out Move Pill Bottle frames, the AUC of ranking a positive frame above the negatives improves by $8.1$ points over the pre-trained policy, compared with $5.1$ points with the trainable projector. The representation also moves further from that of the pre-trained policy (CKA on pre-training frames $0.86$ vs.\ $0.95$), while the contrastive loss stays two to three times higher. A trainable projector therefore absorbs part of the alignment that would otherwise change the representation. These differences do not carry over to success. With either projector, Stage~1 followed by fine-tuning reaches $32.0\%$ ID on Move Pill Bottle and $85.3\%$ vs.\ $87.3\%$ on Place Container, and full RARA reaches $49.3\%$ vs.\ $53.3\%$ and $82.0\%$ vs.\ $84.7\%$ (frozen vs.\ trainable projector, ID). We therefore use the trainable projector by default.

\section{Choice of Pre-training Data}
\label{sec:exp_skillmatch}

This section studies which pre-training data RARA should align to. To test whether the skill shown in the data matters, we swap the pre-training data while keeping everything else fixed: episode count, randomization, cameras and acting arm (Table~\ref{tab:skill_mismatch}). With pick-and-place data, RARA gains $18.8$ and $4.9$ points on the two pick-and-place targets, but only $0.2$ points on Click Alarm Clock. With Click Bell data, the pattern reverses: RARA gains $10.3$ points on Click Alarm Clock, a pressing task like Click Bell, and loses $6.7$ and $20.0$ points on the pick-and-place targets. The anchor therefore tends to help when the pre-training data show the same skill as the target, and it can hurt when the skill differs. This is why RARA retrieves the pre-training data instead of using all of it.

\begin{table}[h]
\centering
\begin{minipage}{0.6\linewidth}
\caption{Gain of RARA over FT (points, RoboTwin) when only the pre-training data is changed. Bold: larger gain for each target.}
\label{tab:skill_mismatch}
{\scriptsize\setlength{\tabcolsep}{3pt}
\begin{tabular*}{\linewidth}{@{\extracolsep{\fill}}lccc@{}}
\toprule
 & \multicolumn{3}{c}{Target task} \\
\cmidrule(l){2-4}
Pre-training data & \shortstack{Move Pill\\Bottle} & \shortstack{Place\\Container} & \shortstack{Click\\Alarm Clock} \\
\midrule
Pick-and-place & $\mathbf{+18.8}$ & $\mathbf{+4.9}$ & $+0.2$ \\
Click Bell & $-6.7$ & $-20.0$ & $\mathbf{+10.3}$ \\
\bottomrule
\end{tabular*}}

\end{minipage}
\end{table}

\section{RETAIN Under a Large Pre-training--Target Gap}
\label{app:retain}

RETAIN~\citep{yadav2025robust} keeps the adapted policy close to the pre-trained one by interpolating their weights, $\theta(\alpha)=(1-\alpha)\theta_{\mathrm{pre}}+\alpha\theta_{\mathrm{ft}}$ (Appx.~\ref{app:robotwin_baselines}). In our experiments, how well this works depends on how far the target task is from the pre-training data. Table~\ref{tab:retain} compares RETAIN with FT in the two simulated settings, and Table~\ref{tab:retain_robocasa} gives the per-task RoboCasa results that are left out of Table~\ref{tab:main}.

\begin{table}[h]
\centering
\caption{RETAIN and FT in the two simulated settings (task-average success rate, \%, mean\,$\pm$\,std).}
\label{tab:retain}
\small
\begin{tabular}{@{}llcccc@{}}
\toprule
 & & \multicolumn{2}{c}{FT} & \multicolumn{2}{c}{RETAIN} \\
\cmidrule(lr){3-4}\cmidrule(lr){5-6}
Setting & Pre-training $\rightarrow$ target & ID & OOD & ID & OOD \\
\midrule
RoboTwin & RoboTwin $\rightarrow$ RoboTwin & 67.4\,$\pm$\,1.4 & 47.3\,$\pm$\,2.4 & 47.0\,$\pm$\,6.7 & 34.2\,$\pm$\,2.3 \\
RoboCasa & DROID (real) $\rightarrow$ RoboCasa (simulated) & 69.2\,$\pm$\,2.5 & 44.3\,$\pm$\,2.0 & 31.1\,$\pm$\,1.8 & 14.8\,$\pm$\,2.9 \\
\bottomrule
\end{tabular}
\end{table}

\begin{table}[h]
\centering
\caption{RETAIN and FT on the five RoboCasa tasks (success rate, \%, mean\,$\pm$\,std).}
\label{tab:retain_robocasa}
\scriptsize
\setlength{\tabcolsep}{2pt}
\resizebox{\linewidth}{!}{%
\begin{tabular}{@{}l*{12}{c}@{}}
\toprule
\textbf{RoboCasa} & \multicolumn{2}{c}{\shortstack{Close Fridge\\Drawer}} & \multicolumn{2}{c}{\shortstack{Open\\Blender Lid}} & \multicolumn{2}{c}{\shortstack{Open\\Cabinet}} & \multicolumn{2}{c}{\shortstack{Turn On\\Toaster}} & \multicolumn{2}{c}{\shortstack{Pick-Place\\to Drawer}} & \multicolumn{2}{c}{\shortstack{Task\\average}} \\
\cmidrule(lr){2-3}\cmidrule(lr){4-5}\cmidrule(lr){6-7}\cmidrule(lr){8-9}\cmidrule(lr){10-11}\cmidrule(lr){12-13}
Method & ID & OOD & ID & OOD & ID & OOD & ID & OOD & ID & OOD & ID & OOD \\
\midrule
FT & 86.0{\fontsize{4}{5}\selectfont$\,\pm$8.7} & 62.7{\fontsize{4}{5}\selectfont$\,\pm$3.1} & 76.0{\fontsize{4}{5}\selectfont$\,\pm$3.5} & 44.7{\fontsize{4}{5}\selectfont$\,\pm$9.0} & 72.0{\fontsize{4}{5}\selectfont$\,\pm$2.0} & 44.0{\fontsize{4}{5}\selectfont$\,\pm$10.6} & 72.0{\fontsize{4}{5}\selectfont$\,\pm$5.3} & 46.0{\fontsize{4}{5}\selectfont$\,\pm$6.0} & 40.0{\fontsize{4}{5}\selectfont$\,\pm$2.0} & 24.0{\fontsize{4}{5}\selectfont$\,\pm$8.0} & 69.2{\fontsize{4}{5}\selectfont$\,\pm$2.5} & 44.3{\fontsize{4}{5}\selectfont$\,\pm$2.0} \\
RETAIN & 33.3{\fontsize{4}{5}\selectfont$\,\pm$5.8} & 22.0{\fontsize{4}{5}\selectfont$\,\pm$6.0} & 13.3{\fontsize{4}{5}\selectfont$\,\pm$7.6} & 9.3{\fontsize{4}{5}\selectfont$\,\pm$9.5} & 46.7{\fontsize{4}{5}\selectfont$\,\pm$3.1} & 19.3{\fontsize{4}{5}\selectfont$\,\pm$4.2} & 46.0{\fontsize{4}{5}\selectfont$\,\pm$2.0} & 20.0{\fontsize{4}{5}\selectfont$\,\pm$2.0} & 16.0{\fontsize{4}{5}\selectfont$\,\pm$3.5} & 3.3{\fontsize{4}{5}\selectfont$\,\pm$3.1} & 31.1{\fontsize{4}{5}\selectfont$\,\pm$1.8} & 14.8{\fontsize{4}{5}\selectfont$\,\pm$2.9} \\
\bottomrule
\end{tabular}%
}
\end{table}

\paragraph{Observation.}
On RoboTwin, the pre-training and target data come from the same simulator, robot and cameras. RETAIN stays below FT on average but keeps about 70\% of its success (47.0\% vs.\ 67.4\% ID, 34.2\% vs.\ 47.3\% OOD), and on Move Pill Bottle it even improves OOD success over FT (13.3\% vs.\ 9.6\%). On RoboCasa, the policy is pre-trained on real-world DROID data and adapted to simulated kitchens. Here RETAIN keeps less than half of the success of FT (31.1\% vs.\ 69.2\% ID, 14.8\% vs.\ 44.3\% OOD), and it falls below FT on all five tasks in both ID and OOD, down to 3.3\% OOD success on Pick-Place to Drawer.

\paragraph{Analysis.}
Weight interpolation tends to work when the pre-trained and fine-tuned weights lie in one connected region of low loss, which is typical when fine-tuning stays close to the pre-training distribution~\citep{wortsman2022robust}. When the target is close to the pre-training data, fine-tuning needs only a small change of the weights, so moving part of the way back toward the pre-trained weights keeps most of the target skill. A likely explanation for the RoboCasa result is that the gap from real-world DROID images to simulated kitchens requires a much larger change of the weights. The interpolated weights then lie between two solutions, and neither the pre-trained nor the fine-tuned behavior is preserved, so both ID and OOD success drop. RARA does not constrain the weights. It constrains the representation of the retrieved demonstrations, so the policy can still change as much as the target task needs, which is consistent with its gains on RoboCasa (Table~\ref{tab:main}).

\end{document}